\documentclass[a4paper,fleqn]{cas-dc}
\PassOptionsToPackage{svgnames,dvipsnames,table,xcdraw}{xcolor}
\usepackage{graphicx}
\newcommand{\fref}[1]{\ref{#1}}
\newcommand{\sref}[1]{Section \ref{#1}}
\newcommand{\tref}[1]{\ref{#1}}
\usepackage{array}
\usepackage{rotating}
\usepackage{amsmath}
\usepackage{mathtools}   
\usepackage[numbers]{natbib}

\usepackage{pifont}
\usepackage{bm}
\usepackage{amssymb}
\usepackage{makecell}  
\usepackage{algorithm}
\usepackage{algpseudocode} 
\usepackage{pdflscape}

\usepackage{array}
\usepackage{multirow}
\usepackage{booktabs}
\usepackage{graphicx}
\usepackage{caption}  
\usepackage[caption=false, font=footnotesize]{subfig}
\usepackage{url}
\usepackage{xcolor}

\usepackage{hyperref}

\AtBeginEnvironment{tabular}{\rmfamily\mdseries} 

\AtBeginEnvironment{table}{\boldmath}   
\AtBeginEnvironment{figure}{}

\begin{document}
\let\WriteBookmarks\relax
\def\floatpagepagefraction{1}
\def\textpagefraction{.001}
\shorttitle{Learning to Sample in Variable Neighborhood Search Algorithm for Urban Cable Routing Optimization}
\shortauthors{Wei Liu et~al.}

\title [mode = title]{Learning to Sample in Variable Neighborhood Search Algorithm for Urban Cable Routing Optimization}

\author[1]{Wei Liu}[orcid=0000-0002-6526-6526]
\ead{weiliu16@nudt.edu.cn}


\affiliation[1]{  
    organization = {National University of Defense Technology},  
    addressline = {109 Deya Road, Kaifu District},   
    city={Changsha},  
    postcode={410073},  
    state={Hunan Province},  
    country={China}  
}  

\author[1]{Rui Wang}[orcid=0000-0001-9048-2979]
\ead{rui_wang@nudt.edu.cn}
\cormark[1]
\cortext[cor1]{Corresponding author}

\author[2]{Chenhui Lin}[orcid=0000-0001-7455-8793]
\affiliation[2]{  
    organization ={Department of Electrical Engineering, Tsinghua University}, 
    addressline ={30 Shuangqing Road},  
    city ={Beijing},  
    postcode ={100084},  
    state ={Beijing Municipality},  
    country ={China}  
}  

\author[1]{Kaiwen Li}[orcid=0000-0003-1550-5987]
\ead{likaiwen@nudt.edu.cn}

\author[1]{Wenhua Li}
\ead{liwenhua1030@aliyun.com}

\author[1]{Tao Zhang}[orcid=0000-0002-0432-2942]
\ead{zhangtao@nudt.edu.cn}

\begin{abstract}
Urban underground cable construction is essential for enhancing power grid reliability, yet the high construction costs demand systematic optimization. Constrained by road network infrastructure, this optimization problem requires consideration not only of connectivity relationships between substations but also of specific routing strategies along road networks, constituting a large-scale bilevel combinatorial optimization problem. Insufficient attention to specific routing subproblems in traditional research and simplistic algorithmic designs that are ill-equipped for large-scale combinatorial optimization leave substantial room for advancement in addressing this complex optimization challenge. To navigate the enormous combinatorial search space, we propose a learning-assisted variable neighborhood search (L-VNS) algorithm integrating four key components. First, an auxiliary task focusing exclusively on the upper-level connectivity subproblem generates high-quality initial solutions by employing hybrid genetic search for connection optimization and A* for detailed path routing. Subsequently, the algorithm iteratively refines the connectivity topology using variable neighborhood search equipped with three complementary operators. A multi-agent deep reinforcement learning module adaptively guides probabilistic neighborhood sampling by jointly encoding both upper-level connectivity patterns and lower-level routing structures, effectively exploiting problem structure. Finally, a modified A* operator re-plans the lower-level paths affected by neighborhood modifications to ensure feasibility and solution completeness. Comprehensive experiments on 12 benchmark instances and 3 GIS-derived instances demonstrate the superiority of L-VNS, achieving total construction cost reductions of 0.92–73.72\% compared to representative approaches. Ablation studies and sensitivity analyses further validate the effectiveness and robustness of the proposed algorithm.
\end{abstract}

\begin{keywords}
Distribution network planning \sep Cable routing \sep Deep reinforcement learning \sep Multiple agents \sep Variable neighborhood search
\end{keywords}

\maketitle

\section{Introduction}

As economic development advances and reliability requirements escalate, an increasing number of cities are planning to comprehensively replace all conventional overhead lines with underground cables in their core urban areas to mitigate the impacts of extreme weather events such as typhoons and ice storms \cite{zhou2024low, LIU2026131837}. However, given the substantial costs associated with this initiative, systematically planning regional cable routing from the ground up in a cost-effective and holistic manner presents a critical optimization challenge \cite{milovsevic2023multi}.

\begin{figure}
	\centering
	\includegraphics[width=1\linewidth]{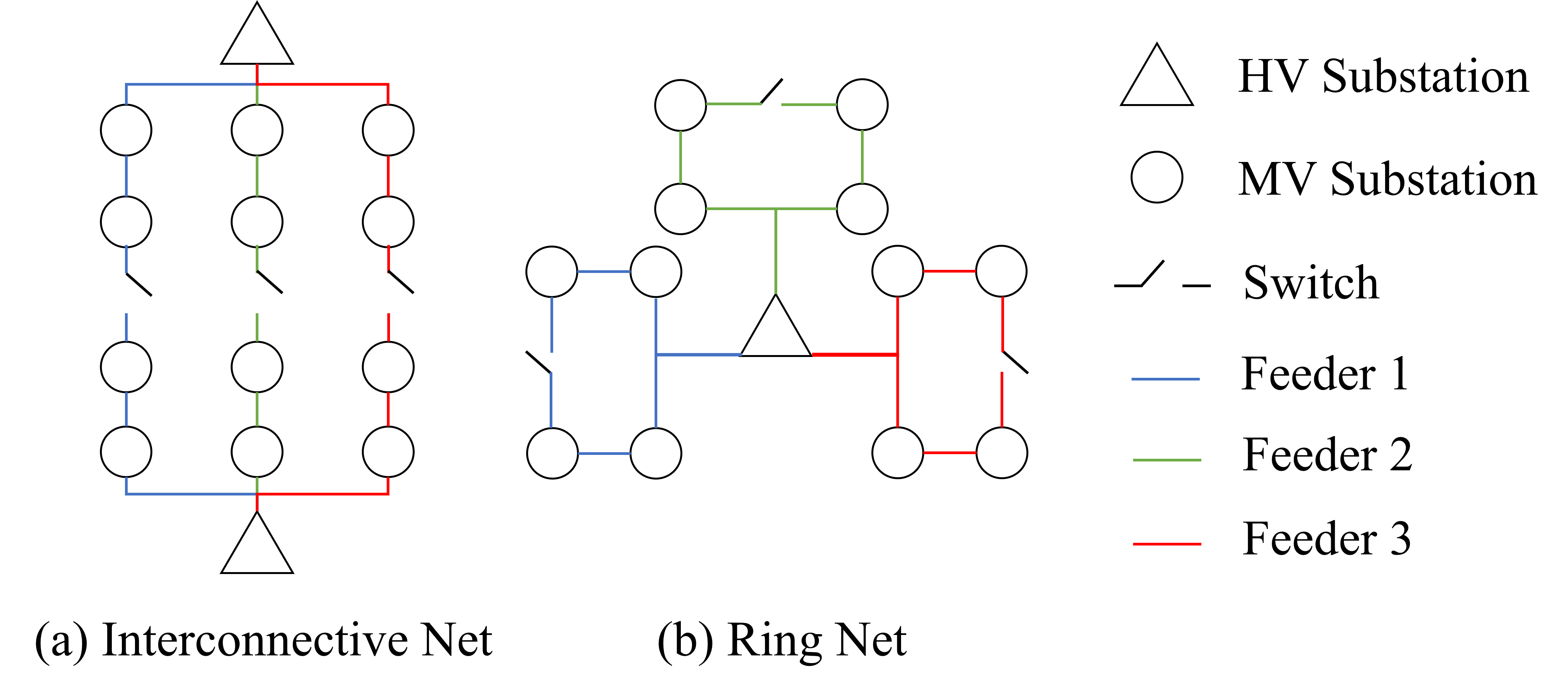}
	\caption{Typical topologies of urban MV distribution cable networks: (a) Interconnected network: MV substations are linked by feeders that originate from one HV substation and terminate at another; (b) Ring network: MV substations are linked by feeders that both originate from and terminate at the same HV substation.}
	\label{fig: cable structure}
\end{figure}

In urban distribution systems, the segment spanning from the 110/10 kV primary substation (i.e., high-voltage (HV) substation) through the 10 kV feeders to the 10/0.4 kV distribution transformers (i.e., medium-voltage (MV) substations) constitutes the MV distribution network (DN). The associated cable infrastructure is typically designed with closed‑loop feeder configurations operated in an open‑loop manner, enabling back‑feeding from multiple sources when required. Common layouts include interconnected and ring feeders, which can be energized from opposite ends or from a common source \cite{diaz2002application}, as illustrated in Fig. \fref{fig: cable structure}. This arrangement supports flexible switching and maintains normal service continuity under any single‑contingency \(N-1\) conditions, thereby enhancing reliability and operational resilience.

\begin{figure}
	\centering
	\includegraphics[width=1\linewidth]{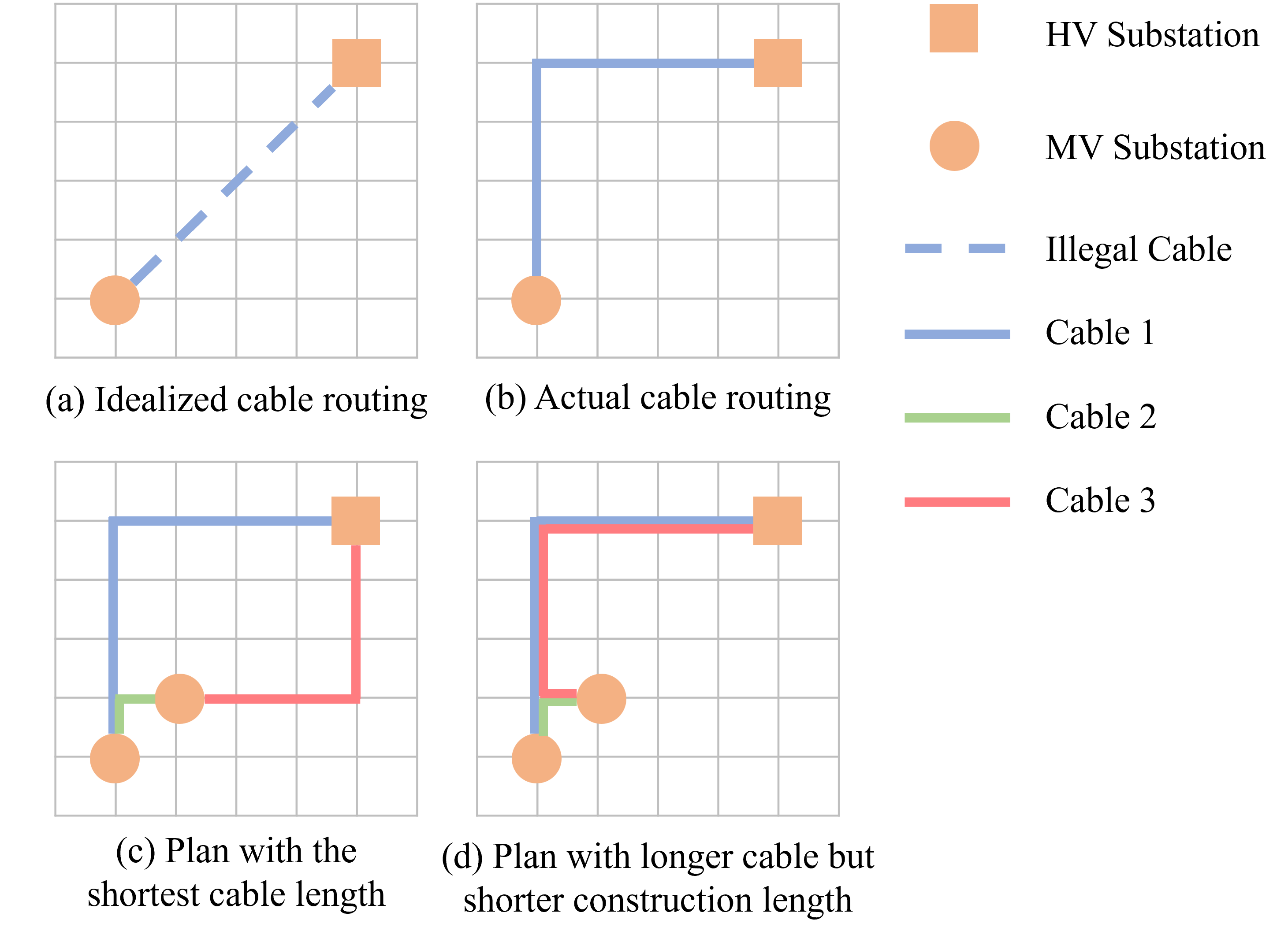}
	\caption{Schematic diagrams of the urban cable routing problem underscore two key points. First, as shown by panel (a) versus panel (b), cables cannot be laid along arbitrary straight-line segments; rather, their paths must adhere strictly to the existing road network topology. Second, as illustrated by panel (c) versus panel (d), planning multiple feeders in parallel along the same road segments can exploit spatial overlap to reduce road-trenching costs, potentially outperforming designs that rely solely on individual shortest-path routes.}
	\label{fig: two statements}
\end{figure}

The cable routing problem is typically formulated as a bilevel combinatorial optimization problem (COP), wherein the objective is to determine the minimum-cost cable routing paths between substations while satisfying capacity constraints, connection requirements, and geographical limitations. The upper-level problem seeks to optimize the interconnection topology among substations, whereas the corresponding lower-level problem determines the detailed cable routing paths given the connectivity configuration decisions. A detailed mathematical formulation of this problem is provided in \sref{Problem Formulation}. In contrast to cable routing challenges in wind farms and rural areas \cite{dutta2012optimal, cazzaro2020heuristic}, urban cables can only be installed beneath roads and not through existing buildings \cite{bai2020hazard}. This restriction complicates decision-making in two key ways: (i) Cable Length Considerations: the feeder length between substations cannot be calculated directly by Euclidean distance, but depends on specific cable routes that strictly along the roads; (ii) Cost Reduction through Overlapping Cables: construction costs include not just the cable procurement cost but also the road-trenching cost, which is much more expensive. Consequently, the shortest-path routing between two substations does not necessarily yield the lowest total construction cost. In many cases, deploying multiple feeders in parallel along the same road segment can reduce the total cost through overlap, even if it increases the aggregate length of cable installed \cite{deveci2019electrical}. Fig. \fref{fig: two statements} gives a schematic diagram to further illustrate these two points. These factors make the problem greatly complex: the decision-making space expands from $2^{N_{s}^2}$ to $2^{(N_{r}+N_{s})^2}$, where $N_{s}$ is the number of substations and $N_{r}$ is the number of road junctions. Typically, \(N_r\) exceeds \(N_s\) by one to two orders of magnitude.

Despite decades of research on the cable routing problem, most studies have focused primarily on the connection relationships between equipment, such as substations or generators. They simplify the consideration of specific cable routes by calculating cable lengths based solely on Euclidean distances between geographic coordinates. Such simplification is typical in less congested environments, as seen in offshore wind farms \cite{cazzaro2020heuristic, cazzaro2023combined, ulku2020optimization, wu2020design, 10970066} and old town districts \cite{bosisio2020optimal}, greatly easing problem solving. However, such simplification is not reasonable in cities, since cables are only allowed to be laid under roads. Several studies further take the geographic information system (GIS) into consideration when optimizing the cable routes \cite{bosisio2018gis, bosisio2021gis, wang2024practical, li2019milp}. The inclusion of road information significantly complicates the problem, making it challenging to address directly with exact planning methods. As a result, dividing the original problem into multiple stages for sequential optimization paired with heuristic approaches has become a common practice \cite{wang2024practical, deveci2019electrical}. Instead of connecting substations directly by a straight cable, the path planning is generally seen as an inner-layer sub-problem or a second-phase sub-problem, where some classical path planning algorithms are applicable, such as the Dijkstra’s algorithm \cite{li2021substation}. A broader set of exact methods, classical heuristics, and artificial intelligence techniques has also been explored, including Benders decomposition \cite{9531373}, genetic algorithms \cite{skok2002genetic, gottlicher2022genetic}, ant colony optimization \cite{liu2016route}, neighborhood search \cite{cazzaro2020heuristic}, reinforcement learning \cite{10694705}, and imitation learning \cite{10388462}.  Furthermore, to address the limited adaptivity in operator selection and parameter configuration of metaheuristic algorithms, recent investigations into learning-assisted neighborhood search have emerged and demonstrated considerable promise in solving COPs \cite{wang2025knowledge, liu2024machine, 9858094}.

Section~\ref{review} provides a detailed review of relevant literature. In summary, despite notable progress in this field, comprehensive research that fully captures the problem structure of urban underground cable routing optimization remains scarce, and existing algorithmic designs for this problem are underdeveloped. On one hand, most existing studies employ overly simplified problem formulations, selectively overlooking the lower-level optimization subproblem (i.e., path routing optimization strongly coupled with urban road networks) while concentrating exclusively on upper-level substation connectivity optimization. On the other hand, current algorithmic designs for urban underground cable routing optimization lack precision. These approaches predominantly rely on multi-stage decomposition strategies with rudimentary heuristic integration, which exhibit excessive greediness and consequently limit optimization accuracy, particularly for large-scale instances. Furthermore, state-of-the-art (SOTA) search algorithms commonly employed in other path planning problems primarily emphasize neighborhood operator design or parameter tuning, while lacking effective search space reduction strategies that are important for addressing such bilevel combinatorial optimization problems characterized by exponentially large search spaces \cite{Ming2025An}.

To address the aforementioned challenges, this work reformulates the urban cable routing problem as a bilevel COP that addresses substation interconnections at the upper level and cable routing paths at the lower level, and constructs a comprehensive set of standard benchmark instances with varying scales for systematic evaluation. From an algorithmic perspective, we propose a learning-assisted variable neighborhood search (L-VNS) algorithm to solve this problem. The algorithm comprises four integrated modules: (i) solution initialization, (ii) a variable neighborhood search (VNS) module dedicated to upper-level optimization, (iii) an $A^*$-based re-planning module for lower-level path routing optimization, and (iv) a deep reinforcement learning (DRL) module for strategic neighborhood sampling in the upper level. The operational workflow proceeds as follows: We first treat the upper-level subproblem as an auxiliary task and apply the hybrid genetic search (HGS) algorithm \cite{VIDAL2022105643} to iteratively optimize and obtain an upper-level-only solution. Subsequently, we employ a modified $A^*$ (hereafter referred to uniformly as $\mathbb{A}^*$) algorithm \cite{hart1968formal} to determine the specific cable routing paths, thereby obtaining a high-quality feasible initial solution. Starting from the initial solution, L-VNS iteratively optimizes substation interconnections using three neighborhood operators: a destruction operator, an intra-feeder 2-opt operator, and an inter-feeder 2-opt operator. During it, the selection of neighborhood operators and the magnitude of destruction scope are adaptively adjusted at each iteration based on historical performance metrics and the degree of stagnation. Apart from that, three pre-trained DRL agents, each corresponding to one of the aforementioned neighborhood operators, compute position-based probability vectors for neighborhood operations conditioned on the current solution state. These probability vectors subsequently guide probabilistic sampling to generate the neighborhood scopes, thereby replacing random sampling with informed strategic selection. Following neighborhood exploration, a modified $\mathbb{A}^*$ operator re-plans the cable routing paths affected by the neighborhood modifications to ensure feasibility and completeness of the resulting solution. Computational experiments demonstrate the advances of the proposed L-VNS: testing on 12 benchmark instances and 3 GIS-derived instances, L-VNS reduces 0.92–73.72\% cost relative to SOTA comparison algorithms. Ablation experiments and sensitive analysis further validate its effectiveness.

The main contributions of this work can be summarized as follows:

\begin{itemize}

\item We formulate the urban cable routing problem as a bilevel optimization framework that decomposes the exponentially large decision space into upper-level connectivity design and lower-level path planning. We additionally provide an open-source standardized benchmark suite for reproducible evaluation.

\item We construct an auxiliary optimization task that relaxes bilevel coupling while preserving essential constraints, and solve it using HGS and $A^*$ to generate high-quality initial solutions that accelerate subsequent search.

\item We develop a bilevel neighborhood search framework where the upper level explores alternative connectivity topologies via variable neighborhood search, and the lower level performs incremental cable path re-planning using a modified trench-sharing-aware $\mathbb{A}^*$ algorithm. By exploiting neighborhood transition relationships, the framework reduces redundant lower-level computations.

\item We integrate a multi-agent DRL module to guide probabilistic neighborhood sampling, replacing uniform random exploration with informed operator and position selection based on learned connectivity and routing patterns.

\end{itemize}

The remainder of this study is organized as follows: \sref{review} surveys related work in detail and concludes our research motivation. \sref{Problem Formulation} formulates the urban cable routing problem. \sref{model} presents the proposed L-VNS method. \sref{sec:settings} illustrates the experimental settings. \sref{results} reports computational experiments on comparison, ablation, and sensitivity.Section~\ref{discussions} presents several high-level discussions of the proposed method and its limitations. \sref{conclusion} concludes this study and outlines future research directions.

\section{Related Work}
\label{review}
\subsection{Studies for Cable Routing Optimization}
\label{Algorithms}

Cable routing optimization is a common task in power systems. Two prominent application domains are photovoltaic plants and wind farms, where the layout of cables connecting photovoltaic (PV) panels and wind turbines is critical for performance and cost efficiency \cite{qiu2024variable, deveci2019electrical, dutta2012optimal, cazzaro2020heuristic, ulku2020optimization}. In these relatively unobstructed environments, planners can prioritize the design of connection relationships among devices and often lay cables along near-linear paths. As a result, the problem is frequently simplified to a minimum spanning tree (MST) formulation, yielding radial network structures that satisfy electrical constraints while minimizing total cable length. A range of classical heuristics and metaheuristics has been applied in this context, including variable-depth large neighborhood search \cite{qiu2024variable}, Dijkstra’s algorithm \cite{dutta2012optimal}, simulated annealing, Tabu search, variable neighborhood search, ant colony optimization, and genetic algorithms \cite{cazzaro2020heuristic}.

However, real-world cable routing faces substantial, often non-negligible obstacles. In outdoor HV transmission, undulating terrain can significantly complicate construction, while in urban MV distribution networks, buildings and other infrastructures may block intended routes. Consequently, it is essential to account not only for connection relationships between devices but also for the specific, feasible routing paths of the cables. 

Monteiro et al. \cite{monteiro2005gis} and Larsson et al. \cite{larsson2025optimizing} addressed point-to-point HV cable routing in realistic geographic settings. The former proposed a dynamic programming approach validated on small-scale case studies, highlighting an iterative and systematic procedure for optimal routing. In contrast, the latter offered a comprehensive comparison of multiple optimization techniques, including Dijkstra’s algorithm, $A^*$ search, and ant colony optimization.

To interconnect urban end-users with substations operating at different voltage levels, cables must be installed strictly along designated roadways. Trageser et al. \cite{trageser2022automated} transformed rendered geographic maps into cost maps and used a greedy strategy to sequence load connections, while employing the $A^*$ algorithm to compute actual route lengths. Wang et al. \cite{wang2024practical} leveraged GIS data to extract street layouts and compute shortest-path distances between substations. To manage complexity, they decomposed the task into two subproblems: closed-loop construction and open-loop operation, solving the former with the Clark–Wright Savings (CWS) heuristic and the latter with a mixed-integer linear programming (MILP) model. Pavon et al. \cite{pavon2019optimal} and Valenzuela et al. \cite{valenzuela2019planning} used GIS data to formulate urban cable routing as a weighted minimum spanning tree and solved it with Prim’s algorithm. \cite{valenzuela2019planning} further applied K-means clustering for simplification. Wang et al. \cite{wang2020practical} represented the road network via a shortest-path adjacency matrix and combined Q-learning with ant colony optimization to form a multi-agent collaborative learning framework. Bosisio et al. \cite{bosisio2020optimal} and Duvnjak et al. \cite{duvnjak2021integrated} simplified routing by optimizing only substation connections, modeling the problem as a mixed-integer program solved with commercial solvers. 

Despite these contributions, research on urban cable routing optimization remains limited in depth, characterized by excessive problem simplification and insufficient treatment of the bilevel coupling between connectivity decisions and routing optimization. Few existing approaches employ advanced algorithmic paradigms capable of handling the complexity of realistic urban cable routing problems.

\subsection{Studies for Learning-assisted Neighborhood Search}
\label{sec:learning-assisted}

Neighborhood search methods have demonstrated outstanding performance in COPs, particularly in path planning problems. Despite the evolution of numerous variants, including Iterated Local Search (ILS), Variable Neighborhood Search (VNS), Large Neighborhood Search (LNS), and Adaptive Large Neighborhood Search (ALNS), one of the core challenges remains how to adaptively manage operator selection, neighborhood scope, and operator parameters during the search process. Confronted with the inherent reliance on manual tuning inherent in metaheuristic algorithms, an increasing body of recent research has been directed toward integrating DRL to enhance algorithm performance~\cite{Ming2025Automated}, yielding promising empirical results.

Many researchers focus on learning-assisted neighborhood operator selection. Lu et al.~\cite{lu2019learning} introduced the ``Learning-to-Improve'' (L2I) framework, wherein a reinforcement learning-based controller iteratively selects improvement operators to refine solutions. These operators are drawn from a robust pool customized for routing problems. Beyond improvement operator selection, Chen et al.~\cite{chen2025learning} further considered perturbation operator selection within neighborhood search and proposed Learning Adaptive Neighborhood Search with Dual operator Selection (LANDS), integrating two reinforcement learning controllers to simultaneously and adaptively select both improvement and perturbation operators. Similarly, Guo et al.~\cite{guo2025learning} constructed a multi-agent convolutional neural network to select perturbation and improvement operators from a low-level heuristic pool, jointly guiding neighborhood search strategies for vehicle routing problems. Wang et al.~\cite{wang2024TETCI} developed a DRL framework built upon the LNS paradigm, wherein the DRL model adaptively selects destruction and repair operators as well as recharging strategies conditioned on the current state of the search process.

On another front, to address the inefficiency of random destruction in LNS, Zhou et al.~\cite{zhou2023learning} employed imitation learning and bipartite graph convolutional networks (GCN) to adaptively select variables for destruction, followed by repair via off-the-shelf solvers such as CPLEX or OR-Tools. Wang et al.~\cite{wang2025adns} simultaneously considered the location and magnitude of destruction operations, training an enhanced graph attention network to output variable-length destruction node sequences. Additionally, several researchers have focused on solution selection among tightly coupled subproblems. For instance, Cao et al.~\cite{cao2024learning} addressed the strong coupling among four subproblems in the integrated port berth allocation and storage problem (IPBP) by introducing a bidirectional long short-term memory (LSTM)-embedded autoencoder network that learns automatically rather than relying on manual design to identify and retain high-quality sub-problem solutions, thereby improving overall solution quality more effectively in global search. More comprehensive discussions on learning-assisted neighborhood search can be found in recent surveys~\cite{szenasi2024machine}.

In summary, while learning-assisted methods provide promising adaptive adjustment mechanisms for neighborhood search variants, however, several critical gaps persist. \textit{First}, most existing studies focus on operator selection or parameter tuning, which does not address the exponentially large search space in bilevel optimization: when neighborhood spaces are vast, efficiently identifying promising solutions within limited sampling budgets imposes stringent demands on model inference performance. \textit{Secondly}, although some researchers have utilized learning methods to autonomously select neighborhood operation positions, there is limited investigation into how bilevel optimization can leverage lower-level solution information during upper-level neighborhood sampling to enhance search efficiency. \textit{Finally}, constructing and training collaborative multi-agent systems for multiple neighborhood operators remains challenging.

\subsection{Research Motivation}

As stated above, existing studies still show gaps in both the problem formulation and algorithm design. To address the aforementioned weaknesses, we tackle the problem from two complementary perspectives.

On the problem formulation level, we formulate the urban cable routing problem as a bilevel COP that jointly optimizes interconnection topology and cable routing paths. This formulation explicitly: (i) preserves the mandated single-loop, multi-source feeder architecture; (ii) restricts all cable segments to the road network; and (iii) internalizes road-trenching costs within the cost function. In parallel, we construct a comprehensive suite of standardized test benchmarks with varying problem scales to enable systematic evaluation of algorithmic performance.

From an algorithmic design perspective, we adopt a learning-assisted search method, namely, L-VNS. First, we develop an auxiliary task to generate high-quality initial solutions, thereby accelerating the overall search process. Subsequently, we devise a VNS method and $\mathbb{A}^*$ algorithm to address upper-level interconnection topology optimization and lower-level cable routing path optimization, respectively. To enhance search efficiency, a multi-agent DRL (MDRL) model is integrated into the upper-level search procedure to autonomously and adaptively sample the neighborhood search space rather than relying on random exploration. The following sections provide detailed descriptions of the problem formulation and the proposed algorithmic framework.

\section{Problem Formulation}
\label{Problem Formulation}
The urban cable routing problem can naturally be formulated as a bilevel optimization problem. At the upper level, the problem optimizes substation connectivity at the substation-graph level, while at the lower level, it determines the routes that realize these connections as feasible paths on the road-graph level. The objective is to minimize the total construction cost, comprising expenses for road-trenching and cable procurement. This section formulates the problem in detail.

\subsection{Notation}
\begin{itemize}
  \item \(S_{hv}\): set of HV substations (depots);
  \item \(S_{mv}\): set of MV substations (customers);
  \item \(V \coloneqq S_{hv} \cup S_{mv}\): set of all substations;
  \item \(\mathcal{F}\): set of feeders, indexed by \(f\); each feeder may start and end at the same depot or at different depots;
  \item \(A \coloneqq \{(i,j)| i\neq j,\ i,j\in V\}\): set of directed arcs between substations;
  \item \(d_{ij}\): distance proxy (e.g., shortest-path length) between substations \(i\) and \(j\);
  \item \(q_i\): contracted load (demand) at MV substation \(i\in S_{mv}\) (MVA);
  \item \(Q\): rated capacity of a feeder (MVA).
  \item \(G=(\mathcal{V},\mathcal{E})\): denotes the directed road graph extracted from GIS data, where $\mathcal{V}$ is the set of road nodes (e.g., intersections, terminals), and $\mathcal{E}$ is the set of directed road segments;
  \item \(\ell_{e}\): geometric length (km) of edge \(e\in\mathcal{E}\);
  \item \(c_{e}^{\mathrm{tr}}\): unit road-trenching cost (million CNY/km) on edge \(e\) (CNY is Chinese Yuan);
  \item \(c_{e}^{\mathrm{ca}}\): unit cable-laying cost (million CNY/km) on edge \(e\);
  \item \(C_{e}^{\max}\): maximum number of parallel cables permitted on edge \(e\).
\end{itemize}

\subsection{Upper Level: Connection Assignment}
\label{sec:stage1}

\paragraph{Decision variable (connectivity among substations):}
\begin{equation}
x_{ij}^{f}\in\{0,1\},
\qquad \forall (i,j)\in A,\ \forall f\in \mathcal{F},
\end{equation}
where \(x_{ij}^{f}=1\) indicates that feeder \(f\) connects substations \(i\) and \(j\).


\paragraph{High-level constraints:}
Because this study does not aim to solve the problem via exact mathematical programming, but to keep the exposition concise and focused on the proposed algorithmic framework, we deliberately omit the full algebraic specification of the numerous and intricate combinatorial constraints. Instead, feasibility is encoded compactly by the abstract set \(\mathcal{X}\):
\begin{equation}
\mathbf{x}\in\mathcal{X},
\end{equation}
which aggregates: (1) \textbf{Network topology constraint:} substation connectivity on the substation graph must conform to a ring or interconnected topology, as illustrated in Fig. \fref{fig: cable structure}; and (2) \textbf{Feeder capacity constraint:} for every feeder \(f\in\mathcal{F}\), the aggregate demand of MV substations assigned to \(f\) must not exceed the feeder capacity \(Q\).

Define the set of abstract connections retained in the upper level as
\begin{equation}
\mathcal{A}(\mathbf{x})\coloneqq\{(i,j,f)\in V\times V\times \mathcal{F} | x_{ij}^{f}=1\},
\end{equation}
it constitutes the input pairs for the lower-level subproblem.

\subsection{Lower Level: Route Refinement}

The lower level implements the abstract connections \(\mathcal{A}(\mathbf{x})\) on the road graph by choosing which edges to excavate and how many parallel cables to place on each.

\paragraph{Decision variables (connectivity on the road graph):}
\begin{equation}
y_{e}\in\{0,1\},\qquad
k_{e}\in\{0,1,\ldots, C_{e}^{\max}\},
\qquad \forall e\in\mathcal{E},
\end{equation}
where \(y_e = 1\) denotes that edge \(e\) is excavated (used), and \(k_e\) is the number of parallel cables laid on \(e\).

\paragraph{Objective function:}
\begin{equation}
\min F = \sum_{e\in\mathcal{E}}
\Bigl(c_{e}^{\mathrm{tr}}\,\ell_{e}\,y_{e}
      +c_{e}^{\mathrm{ca}}\,\ell_{e}\,k_{e}\Bigr).
  \label{eq: objective}
\end{equation}

The objective considers both cable procurement and road-trenching costs.

\paragraph{High-level constraints:}

feasibility is encoded by a set \(\mathcal{Y}(\mathcal{A}(\mathbf{x}))\):
\begin{equation}
(\mathbf{y},\mathbf{k})\in \mathcal{Y}\bigl(\mathcal{A}(\mathbf{x})\bigr),
\end{equation}
which encompasses: (1) \textbf{Connectivity realization:} for every abstract link \((i,j,f)\in\mathcal{A}(\mathbf{x})\), the nodes associated with substations \(i\) and \(j\) are connected by a path in the subgraph induced by \(\{e\in\mathcal{E}: y_{e}=1\}\); (2) \textbf{Trenching-cabling consistency:} cables are laid only on excavated edges (trenching can be shared across multiple cables).

\subsection{Integrated Formulation}
The problem can be cast as a bi-level structure that minimizes the construction cost \(F\) while coupling connectivity and routing:
\begin{equation}
\begin{aligned}
\min_{\mathbf{x} \in \mathcal{X}} \quad & F(\mathbf{x}, \mathbf{y}^*, \mathbf{k}^*) \\
\text{s.t.} \quad & \mathbf{x} \in \mathcal{X}, \\
& (\mathbf{y}^*, \mathbf{k}^*) \in \arg\min_{\mathbf{y} \in \mathbf{Y}, \mathbf{k} \in \mathbf{K}} \\
& \quad \quad \quad \quad \left\{ F(\mathbf{y}, \mathbf{k} | \mathbf{x}) : (\mathbf{y}, \mathbf{k}) \in \mathcal{Y}\bigl(\mathcal{A}(\mathbf{x})\bigr) \right\}.
\end{aligned}
\end{equation}

Given \(|\mathcal{F}|\) feeders, \(|A|\) candidate substation arcs, and \(|\mathcal{E}|\) road edges, there are \(|\mathcal{F}||A| + 2|\mathcal{E}|\) binary variables, often surpassing \(10^{6}\) for typical urban settings. The resulting formulation is tightly constrained by structural and capacity requirements, rendering it a large-scale, multi-constraint combinatorial optimization problem that lies beyond the practical capabilities of standard exact MIP solvers.

For clarity, we denote the upper- and lower-level decision variables as $x_u$ and $x_l$, respectively, in the following text.

\section{Proposed Method}
\label{model}
Fig. \fref{fig: framework} presents the overall architecture of the proposed L‑VNS algorithm, which consists of (1) solution initialization, (2) upper-level variable neighborhood search (U-VNS), (3) multi-agent DRL for guided sampling, (4) lower-level re-planning (L-Replan) with $\mathbb{A}^*$, and (5) solution evaluation and update. Algorithm~\ref{alg:lvns} presents the detailed pseudocode. The algorithm consists of four integrated phases executed iteratively. First, an initial feasible solution $(\mathbf{x}_u, \mathbf{x}_l)$ is obtained by reformulating the upper-level connectivity problem as a multi-depot capacitated vehicle routing problem (MD-CVRP) and solving it via HGS, followed by $\mathbb{A}^*$ to compute cable routing paths on the road graph $G_r$; the operator weight vector $w$ is initialized uniformly (Lines 1-2). In each iteration, the U-VNS module generates $N$ candidate upper-level solutions $\{\tilde{\mathbf{x}_u}_i\}_{i=1}^N$ through a combination of random perturbation and DRL-guided action sampling (Line 4). For each candidate solution, the L-Replan module re-plans affected cable paths, obtaining the corresponding lower-level solutions $\{\tilde{\mathbf{x}_l}_i\}_{i=1}^N$ and their costs $\{\tilde{F}^{(i)}\}_{i=1}^N$ (Line 5). The best candidate solution $(\hat{\mathbf{x}_u}, \hat{\mathbf{x}_l}, \hat{F})$ is selected, and if it improves upon the incumbent, both the solution $\mathbf{x}_u^\star, \mathbf{x}_l^\star$ and the best cost $F^\star$ are updated (Lines 6-9). Finally, operator weights $w$ are adaptively adjusted to reflect successful and unsuccessful decisions (Line 10), and this process repeats until the termination criterion (maximum running time) is satisfied. The detailed mechanisms of each component, including solution initialization, the upper-level search, the DRL-guided sampling, and the lower-level re-plan, are elaborated in the following subsections.

Such a design aims to address the core computational challenge in bilevel optimization \cite{huang2023bilevel}: each upper-level candidate solution necessitates a complete lower-level optimization procedure, resulting in prohibitively expensive computational costs. We mitigate this challenge through four synergistic strategies. First, we construct an auxiliary MD-CVRP task to obtain a high-quality initial solution via HGS and a single $A^*$ pass, circumventing extensive bilevel evaluations during early exploration. Second, neighborhood search operators (unlike evolutionary algorithms that drastically modify solution structures) impose structural locality through localized perturbations, enabling efficient reuse of lower-level routing information from parent solutions. Third, we implement incremental re-planning where only affected cable routes are re-optimized via $\mathbb{A}^*$ algorithm while unaffected routes are directly inherited, with edge costs dynamically adjusted to capture trench-sharing benefits. Fourth, a multi-agent DRL module learns to predict high-quality perturbation positions by encoding both upper-level connectivity and lower-level routing information, concentrating search effort on promising neighborhoods. Collectively, these strategies avoid extensive early exploration, leverage structural locality, selectively re-optimize affected components, and prioritize high-potential neighborhoods, thereby enabling efficient navigation of the bilevel search space.

\begin{figure*}
	\centering
	\includegraphics[width=1\linewidth]{framework-detail.jpg}
	\caption{
	Workflow of the proposed L-VNS algorithm. The framework comprises:	\textbf{(1) Solution Initialization} using HGS-based MD-CVRP auxiliary task followed by standard $A^*$ routing to generate a feasible seed solution. \textbf{(2) Upper-Level Variable Neighborhood Search} with three stages: (i) uniform random perturbation, (ii) DRL-guided probability sampling, and (iii) neighborhood generation via three operators. \textbf{(3) Multi-Agent DRL Module} generating edge-specific probability vectors through LSTM (sequential dependency), multi-head attention (cross-feeder interaction), and fully connected (FC) layers. \textbf{(4) Lower-Level Re-Planning} with $\mathbb{A}^*$ that dynamically adjusts edge costs based on accumulated cable routes to capture shared trenching benefits. \textbf{(5) Solution Evaluation and Update} with adaptive operator weight adjustment. On the right side of the figure, a simplified core workflow and solution schematic are provided.}
	\label{fig: framework}
\end{figure*}

\begin{algorithm}[htb]
\caption{Learning-assisted Variable Neighborhood Search (L-VNS)}
\label{alg:lvns}
\begin{algorithmic}[1]
\Require Upper-level graph $G_u$, road graph $G_r$, DRL agents $\mathcal{\pi} =\{\pi_1, \pi_2, \pi_3\}$,
         neighborhood operators $\mathcal{O} = \{\mathbb{O}_1, \mathbb{O}_2, \mathbb{O}_3\}$
\Ensure Best solution $(\mathbf{x}_u^\star, \mathbf{x}_l^\star)$ with cost $F^\star$
\Statex \textit{\% Solution Initialization}
\State $(\mathbf{x}_u, \mathbf{x}_l) \leftarrow \text{HGS}(G_u) + \mathbb{A}^*(G_r)$
\State $\mathbf{x}_u^\star \leftarrow \mathbf{x}_u$, $F^\star \leftarrow F(\mathbf{x}_u, \mathbf{x}_l)$, $w \leftarrow [1, 1, 1]$
\While{not stopping criteria}
  \Statex \quad \textit{\% Upper-Level Search}
  \State $\{\tilde{\mathbf{x}_u}_i\}_{i=1}^N \leftarrow \text{U-VNS}(\mathbf{x}_u^\star, \mathbf{x}_l^\star, \mathcal{O}, \mathcal{\pi}, w)    $
  \Statex \quad \textit{\% Lower-Level Re-Plan and Evaluation}
\State $\{\tilde{\mathbf{x}_l}_i, \tilde{F}^{(i)}\}_{i=1}^N \leftarrow \text{L-Replan}(\{\tilde{\mathbf{x}_u}_i\}_{i=1}^N)$

    \Statex \quad \textit{\% Update solutions}
  \State $(\hat{\mathbf{x}_u}, \hat{\mathbf{x}_l}, \hat{F}) \leftarrow \arg\min_i \tilde{F}^{(i)}$
  \If{$\hat{F} < F^\star$}
    \State $\mathbf{x}^\star \leftarrow \hat{\mathbf{x}}$, $F^\star \leftarrow \hat{F}$
  \EndIf
    \State $w \leftarrow \text{UpdateW}(w)$

\EndWhile
\State \textbf{return} $\mathbf{x}_u^\star, \mathbf{x}_l^\star, F^\star$
\end{algorithmic}
\end{algorithm}

\subsection{Auxiliary Task Construction for Solution Initialization}
\label{Auxiliary}

We formulate an auxiliary optimization problem that relaxes the original bilevel problem by focusing on upper-level connectivity while replacing explicit cable-routing decisions with inter-substation shortest-path distances:

\begin{equation}
\min F'= \sum_{f\in \mathcal{F}}\sum_{(i,j)\in A} d_{ij}\,x_{ij}^{f}.
\end{equation}

This reformulation transforms the problem into a Multi-Depot Capacitated Vehicle Routing Problem (MD-CVRP) over the substation set, where road-graph shortest-path distances serve as arc costs. Since cable cost scales proportionally with cable length, connectivity patterns minimizing shortest-path distances typically also yield favorable cable-routing costs. While the trench-sharing objective may occasionally trade increased cable length for reduced excavation, such trade-offs typically remain close to the ``minimum-cable-length'' optimum. Therefore, the relaxed solution provides a strong ``anchor'' in the connectivity space that is proximate to the bilevel optimum. Moreover, the relaxation enables computationally efficient initialization, which is critical for bilevel optimization where exact lower-level evaluations are expensive. Although this auxiliary solution does not capture trench-sharing benefits from joint trenching, it furnishes a high-quality feasible starting point that subsequent neighborhood search refines by exploiting these opportunities.

This auxiliary optimization problem is solved through the HGS algorithm \cite{VIDAL2022105643}, a state-of-the-art metaheuristic for vehicle-routing variants. HGS efficiently explores the solution landscape by combining evolutionary operators with local search techniques. Given the resulting upper-level solution (feeder-level connection pairs), we compute the corresponding lower-level solution (point-to-point routes) via the $A^*$ algorithm \cite{hart1968formal} on the road graph, ensuring optimal paths under the chosen edge metrics.

\subsection{Upper-Level Variable Neighborhood Search}
\label{sec:mvns}

Algorithm~\ref{alg:U-VNS} details the upper-level neighborhood search procedure. The algorithm operates in three integrated phases to generate $N$ candidate solutions ($N=30$ in default). In the shake phase (Lines 1-3), a destruction operator is uniformly randomly selected from the operator set $\mathcal{O}$ and applied to the current best solution $\mathbf{x}_u^\star$ to generate an initial perturbed solution; the corresponding lower-level cable routes are re-planned via $\mathbb{A}^*$ to establish the basis for feature extraction. In the sampling phase with DRL guidance (Lines 4-7), an improvement operator is selected via categorical distribution weighted by the operator weight vector $w$, ensuring that high-performing operators are favored while maintaining exploration diversity. Solution features $\mathbf{s}$ are extracted from both the perturbed upper-level structure and the re-planned lower-level paths; a pre-trained DRL agent paired with the selected improvement operator computes a position-aware probability vector $\mathbf{p}$ that guides strategic action sampling. Finally, in the neighborhood generation phase (Lines 8-10), the $N$ sampled actions are applied sequentially to the perturbed solution via the improvement operator, yielding $N$ distinct candidate neighborhood solutions. This DRL-guided sampling mechanism replaces uniform random selection, leveraging learned patterns to prioritize promising search directions while maintaining computational efficiency. The detailed design of individual operators and the DRL agent architecture are discussed in subsequent sections.

\begin{algorithm}[htb]
\caption{Upper-Level Variable Neighborhood Search (U-VNS)}
\label{alg:U-VNS}
\begin{algorithmic}[1]
\Require Current solution $\mathbf{x}_u^\star, \mathbf{x}_l^\star$, operators $\mathcal{O} = \{\mathbb{O}_1, \mathbb{O}_2, \mathbb{O}_3\}$,
         DRL agents $\mathcal{\pi} = \{\pi_1, \pi_2, \pi_3\}$, operator weights $w$, 
         neighborhood size $N$
\Ensure Candidate neighborhood solutions $\{\tilde{\mathbf{x}_u}_1, \ldots, \tilde{\mathbf{x}_u}_N\}$
\Statex \textit{\% Phase 1: Shake}
\State $o_{\text{shake}} \sim \text{Uniform}(\mathcal{O})$ 
\State $\tilde{\mathbf{x}_u} \leftarrow o_{\text{shake}}(\mathbf{x}_u^\star)$ 
\State $\tilde{\mathbf{x}_l} \leftarrow \mathbb{A}^*(\tilde{\mathbf{x}_u}, G_r)$

\Statex \textit{\% Phase 2: Sampling with DRL Guidance}
\State $o_{\text{improve}} \sim \text{Categorical}(w)$ 
\State $\mathbf{s} \leftarrow \Phi(\tilde{\mathbf{x}_u}, \tilde{\mathbf{x}_l})$ 
\State $\mathbf{p} \leftarrow \pi_{o_{\text{improve}}}(\mathbf{s})$ 
\State Sample $N$ actions: $\{\mathbf{a}_1, \ldots, \mathbf{a}_N\} \sim \mathbf{p}$ 
\Statex \textit{\% Phase 3: Generate Neighborhoods}
\For{$i = 1$ to $N$}
  \State $\tilde{\mathbf{x}_u}_i \leftarrow o_{\text{improve}}(\tilde{\mathbf{x}_u}^{(0)}, \mathbf{a}_i)$ 
\EndFor
\State \textbf{return} $\{\tilde{\mathbf{x}_u}_1, \ldots, \tilde{\mathbf{x}_u}_N\}$
\end{algorithmic}
\end{algorithm}

\begin{figure}
	\centering
	\includegraphics[width=0.6\linewidth]{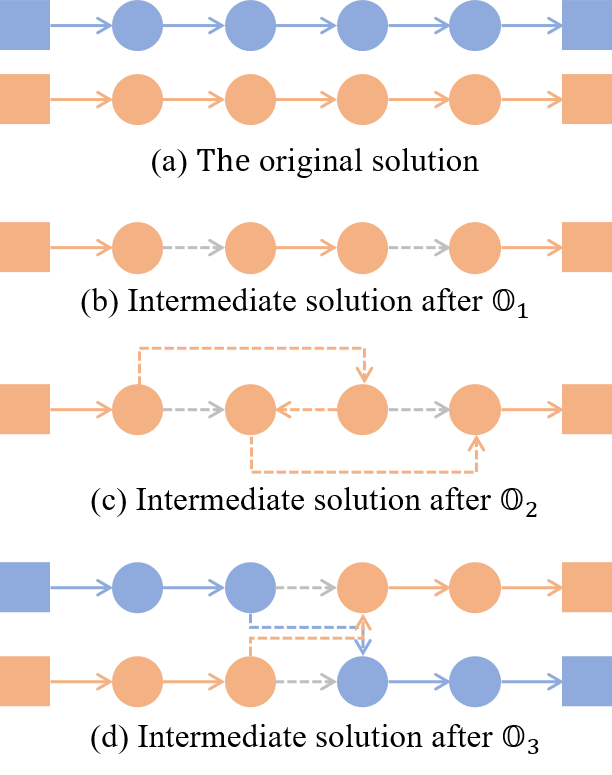}
	\caption{The operation diagrams for the three destruction operators: squares indicate HV substations, circles denote MV substations, orange and blue modules signify two independent feeders, gray dotted lines represent removed segments, and orange and blue dotted lines represent newly connected segments.}
	\label{fig: destruction operators}
\end{figure}

Fig. \fref{fig: destruction operators} illustrates the three upper-level search operators embedded in the U-VNS module. Each operator, paired with a pre-trained DRL agent, serves as an improvement operator to generate diversified neighborhoods. The operators modify the current substation connectivity structure without determining the specific cable routing.

\begin{itemize}

\item $\mathbb{O}_1$ (path-destruction): Randomly removes connections between $\kappa(st)$ pairs of interconnected substations, with the perturbation magnitude adjusted based on stagnation iterations (Eq. \eqref{eq: delta}). Initially, when stagnation is low, fewer connections (2–4 pairs) are removed for fine local refinement; as stagnation increases, the number of removed connections rises (6–8 pairs) to enhance large-scale diversification and escape local optima. This adaptive approach effectively balances exploitation and exploration by increasing perturbation strength during extended stagnation.

\begin{equation}
\kappa(st) = 
\begin{cases} 
2, & \text{if } st < 20, \\ 
4, & \text{if } 20 \leq st < 30, \\ 
6, & \text{if } 30 \leq st < 40, \\ 
8, & \text{if } st \geq 40.
\end{cases}
\label{eq: delta}
\end{equation}

\item $\mathbb{O}_2$ (intra-feeder 2-opt): Selects two non-adjacent edges along the same feeder's substation sequence and performs 2-opt recombination by reversing the intervening segment. This operator refines the substation ordering within individual feeders to reduce inter-substation distances.

\item $\mathbb{O}_3$ (inter-feeder 2-opt): Selects one edge from each of two distinct feeders and exchanges the corresponding substation segments. This operator redistributes terminal stations across feeders to balance feeder loads and improve connectivity.

\end{itemize}

\subsection{Multi-Agent Deep Reinforcement Learning for Guided Sampling}
\label{sec:DRL4sample}

To address the scalability challenges of traditional VNS on large-scale instances, each destruction operator is paired with a pre-trained DRL agent $\pi_1, \pi_2, \pi_3$ that learns to predict edge-specific destruction probabilities conditioned on the current solution state. Rather than uniform random sampling, the DRL agent ingests solution features and outputs a probability vector for edge removal; weighted sampling based on this vector constructs candidate neighborhoods more efficiently. The above process corresponds to lines 5-7 in Algorithm~\ref{alg:U-VNS}. To preserve exploration and mitigate premature convergence, neighborhoods are generated via DRL-guided sampling with probability 0.7 and via uniform random selection with probability 0.3.

\subsubsection{State Representation}

The state $s_t$ encodes the current best solution at iteration $t$ as a three-dimensional tensor of shape $E \times M \times D$, where $E$ is the number of substation-to-substation paths, $M$ is the maximum number of nodes per path (including road nodes), and $D = 2$ stores the planar coordinates of nodes. This compact tensor representation captures the spatial structure of the network and facilitates efficient neural network processing. Critically, by incorporating the concrete cable routing information from the incumbent lower-level solution as prior knowledge, the state representation establishes a direct connection between the upper-level sampling decisions and lower-level route configurations. This alignment between state features and the reward signal, which evaluates the objective improvement of newly sampled solutions, enables the DRL agent to learn effective exploration strategies that respect the interdependencies between upper and lower decision levels, rather than treating them independently.

\subsubsection{Neural Network Architecture and Forward Computation}

\begin{figure*}
	\centering
	\includegraphics[width=1\linewidth]{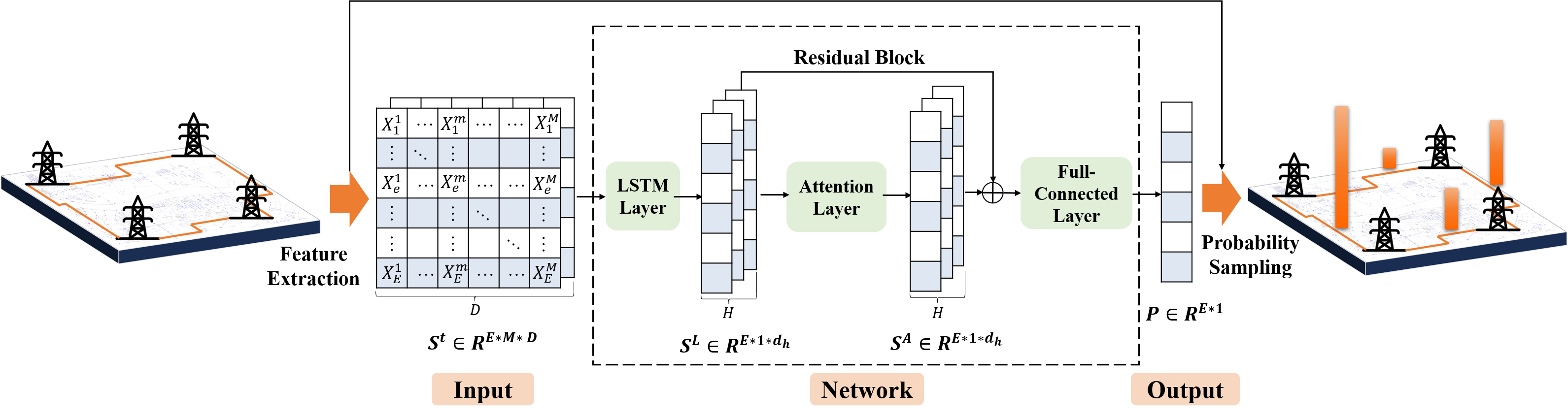}
	\caption{Schematic of the neural network architecture and feedforward computation. Each agent comprises a sequential stack of an LSTM layer, a multi-head attention layer, and a fully connected layer. The network ingests a three-dimensional tensor that encodes the two-dimensional coordinates of all nodes along the routes in the current best solution. The output is a probability vector over existing edges, indicating their likelihood of being sampled. By repeatedly performing probability sampling from this probability vector, the model generates a set of structurally perturbed upper-level neighborhood solutions.}
	\label{fig: network}
\end{figure*}

As illustrated in Fig. \fref{fig: network}, each DRL agent comprises three sequential components: (i) an LSTM module to capture sequential dependencies within individual cable feeders; (ii) a multi-head attention module to model spatial interactions across different feeders; and (iii) a two-layer fully connected module for regression that outputs a probability vector over existing edges. The outputs of the LSTM and attention modules are combined via a residual connection to preserve sequential features and stabilize training. Each DRL agent maps the state $s_t$ to an action probability vector $p$, as follows:

\textbf{LSTM encoding.} Given the state tensor $s_t \in \mathbb{R}^{E \times M \times 2}$ at iteration $t$, the LSTM processes each edge sequence independently:
\begin{equation}
\mathbf{H}_{\text{lstm}} = \text{LSTM}(s_t; \theta_{\text{lstm}}), \quad \mathbf{H}_{\text{lstm}} \in \mathbb{R}^{E \times M \times d_h},
\end{equation}
where $d_h = 64$ is the hidden dimension. We extract the final hidden state for each edge:
\begin{equation}
\mathbf{h}_i = \mathbf{H}_{\text{lstm}}[i, -1, :], \quad i = 1, \ldots, E,
\end{equation}
yielding $\mathbf{H} = [\mathbf{h}_1, \ldots, \mathbf{h}_E]^\top \in \mathbb{R}^{E \times d_h}$.

\textbf{Multi-head attention.} To model inter-edge dependencies, we apply self-attention over the edge representations:
\begin{equation}
\mathbf{A} = \text{MultiheadAttn}(\mathbf{H}, \mathbf{H}, \mathbf{H}; \theta_{\text{attn}}) \in \mathbb{R}^{E \times d_h}.
\end{equation}

\textbf{Residual connection.} The LSTM and attention outputs are combined via element-wise addition:
\begin{equation}
\mathbf{Z} = \mathbf{H} + \mathbf{A} \in \mathbb{R}^{E \times d_h}.
\end{equation}

\textbf{Probability vector generation.} A two-layer fully connected network maps the combined representation to edge-specific logits, which are normalized via softmax:
\begin{equation}
\mathbf{l} = \text{ReLU}\bigl(\mathbf{Z} \mathbf{W}_1 + \mathbf{b}_1\bigr) \mathbf{W}_2 + \mathbf{b}_2, \quad \mathbf{l} \in \mathbb{R}^{E},
\label{eq:fc_layers}
\end{equation}

\begin{equation}
\tilde{\mathbf{l}} = \text{clip}(\mathbf{l}, -10, 10), 
\end{equation}

\begin{equation}
p(a_i | s_t; \theta) = \frac{\exp(\tilde{l}_i)}{\sum_{j=1}^{E} \exp(\tilde{l}_j)}, \quad i = 1, \ldots, E,
\end{equation}
where $\theta_{\text{lstm}}$, $\theta_{\text{attn}}$, $\mathbf{W}_1 \in \mathbb{R}^{d_h \times d_h}$,   $\mathbf{W}_2 \in \mathbb{R}^{d_h \times 1}$, $\mathbf{b}_1$, and $\mathbf{b}_2$ are learnable parameters, and the clipping operation stabilizes training by preventing extreme logit values. The resulting probability vector $\mathbf{p} = [p(a_1 | s_t; \theta), \ldots, p(a_E | s_t; \theta)]^\top$ indicates the likelihood of each edge being selected for destruction.

\subsubsection{Operator-Specific Action Sampling}  
\label{sec:sampling}

Based on the calculated probability vector, the three DRL agents employ distinct sampling strategies tailored to their respective operators:

\begin{itemize}

\item $\pi_1$ (path-destruction): The agent performs weighted sampling without replacement to select exactly $\kappa(st)$ edges (paths) for destruction, where $\kappa(st)$ is determined by the stagnation time as in Eq. \eqref{eq: delta}.

\item $\pi_2$ (intra-feeder 2-opt): The agent samples two edges within the same feeder by first selecting a feeder according to its aggregated edge probabilities, then sampling two non-adjacent edges within that feeder according to their normalized probabilities.

\item $\pi_3$ (inter-feeder 2-opt): The agent first samples two distinct feeders, then samples one edge from each feeder according to their respective probabilities within the probability vector $p(a | s_t)$.

\end{itemize}

By repeated sampling from these probability distributions, the model generates a set of $N$ structurally perturbed neighborhood solutions, each serving as a candidate for evaluation in the upper-level optimization loop.

\subsubsection{Reinforcement Learning}

The DRL agent is trained offline using the REINFORCE policy gradient algorithm on a designated training distribution $\mathcal{D}$, which generates diverse feasible initial solutions for small-scale instances (e.g., $(n_{grid}, N_{\text{MV}}, N_{\text{HV}}) = (20, 30, 5)$ as stated in \sref{sec:benchmark}). Each search operator $\mathbb{O}_k$ ($k \in \{1,2,3\}$) maintains its own policy network $\pi_k(\cdot; \theta)$. Algorithm~\ref{alg:drl-training} outlines the complete training procedure.

\begin{algorithm}[htb]
\caption{REINFORCE-Based DRL Agent Training}
\label{alg:drl-training}
\begin{algorithmic}[1]
\Require Training distribution $\mathcal{D}$, target operator $\mathbb{O}_k$,
         max epochs $E_{\max}=1000$, samples $N_{\text{sample}}=500$, stagnation limit $L = 20$
\Ensure Trained policy $\pi_k(\cdot; \theta)$
\State Randomly initialize policy network $\theta$
\For{$epoch = 1$ to $E_{\max}$}
  \State Initialize experience buffer: $\mathcal{B} \leftarrow \emptyset$
  
  \For{$n = 1$ to $N_{\text{sample}}$} 
    \State Generate random feasible initial solution $s_0 \sim \mathcal{D}$
    \State $s \leftarrow s_0$, $F^* \leftarrow F(s)$, $c \leftarrow 0$
    
    \While{$c < L$}
      \State Extract state: $\mathbf{s}_t \leftarrow \Phi(s)$
      \State Compute probability: $\mathbf{p}_t \leftarrow \pi_k(\mathbf{s}_t; \theta)$
      \State Sample action: $a_t \sim \mathbf{p}_t$
      \State Apply and re-plan: $s' \leftarrow \mathbb{A}^*(\mathbb{O}_k(s, a_t), G_r)$
      \State Compute reward: $r_t \leftarrow$~Eq. \eqref{eq: reward}
      \If{$r_t > 0$}
        \State $\mathcal{B} \leftarrow \mathcal{B} \cup \{(\mathbf{s}_t, a_t, r_t)\}$
        \State $s \leftarrow s'$, $F^* \leftarrow F(s')$, $c \leftarrow 0$
      \Else
        \State $c \leftarrow c + 1$
      \EndIf
    \EndWhile
  \EndFor
  
  \If{$|\mathcal{B}| > 0$}
    \State Update $\theta$ as Eqs.~\eqref{eq: loss}-\eqref{eq: theta}
  \EndIf
\EndFor

\State \textbf{return} $\pi_k(\cdot; \theta)$
\end{algorithmic}
\end{algorithm}

\textbf{Training Strategy.} 
For each training epoch, we generate $N_{\text{sample}}=500$ random feasible initial solutions from $\mathcal{D}$. Starting from each initial solution $s_0$, the agent iteratively applies the target operator $\mathbb{O}_k$ guided by its current policy $\pi_k(\cdot; \theta)$ until reaching a stagnation limit of $L=20$ consecutive non-improving iterations. This iterative search mimics the actual deployment scenario where the operator is repeatedly invoked within the VNS framework. Critically, only experiences yielding positive improvements (i.e., $r_t > 0$) are stored in the experience buffer $\mathcal{B}$ to focus learning on successful sampling strategies and avoid diluting the gradient signal with uninformative zero-reward transitions.

\textbf{Reward Design.}
After applying the upper-level search operator and re-planning via $\mathbb{A}^*$ algorithm to obtain the new solution with cost $F(s')$, the reward is computed as the relative improvement over the best-so-far solution:
\begin{equation}
r_t = 
\begin{cases} 
\dfrac{F^* - F(s')}{F^*}, & \text{if } F(s') < F^*, \\
0, & \text{otherwise},
\end{cases}
\label{eq: reward}
\end{equation}
where $F^*$ denotes the incumbent best cost. This normalized reward formulation ensures scale-invariance across instances of different magnitudes and provides a consistent learning signal.

\textbf{Policy Gradient Update.}
At the end of each epoch, the accumulated experience buffer $\mathcal{B} = \{(\mathbf{s}_i, \mathbf{a}_i, r_i)\}_{i=1}^{|\mathcal{B}|}$ contains only successful improvement transitions, where $\mathbf{a}_i$ denotes the set of sampled actions (e.g., two edge indices for 2-opt operators). The policy network parameters $\theta = \{\theta_{\text{lstm}}, \theta_{\text{attn}}, \mathbf{W}_1, \mathbf{b}_1, \mathbf{W}_2, \mathbf{b}_2\}$ are updated via stochastic gradient descent on mini-batches sampled from $\mathcal{B}$ using the REINFORCE objective:
\begin{equation}
L(\theta) = -\frac{1}{|\mathcal{B}|} \sum_{i=1}^{|\mathcal{B}|} r_i \log \pi_k(\mathbf{a}_i | \mathbf{s}_i; \theta),
\label{eq: loss}
\end{equation}
where $\pi_k(\mathbf{a}_i | \mathbf{s}_i; \theta)$ represents the joint probability of the action set $\mathbf{a}_i$ under the sampling strategy described in Section~\ref{sec:sampling}. The gradient directly weights the log-probability of each action set by its observed reward. The parameter update follows:
\begin{equation}
\theta \gets \theta - \alpha \nabla_\theta L(\theta),
\label{eq: theta}
\end{equation}
with learning rate $\alpha$. The gradient is computed via automatic differentiation and backpropagated through the entire network architecture. This training process repeats for $E_{\max}=1000$ epochs to ensure convergence. The parameters as for the DRL model are tuned using IRACE \cite{lopez2016irace}. The parameter combination with the best performance is reported in Table~\tref{tab:drl-parameters}.

For each test instance, the pre-trained agents undergo a brief online fine-tuning phase using the search trajectories from the first 100 iterations of L-VNS deployment. This transfer learning mechanism enables rapid adaptation to instance-specific characteristics, thereby accelerating convergence during online deployment.

\begin{table}[htb]
\centering
\caption{Parameter settings for DRL agent construction and training.}
\label{tab:drl-parameters}
\begin{tabular}{lc}
\toprule
\textbf{Parameter} & \textbf{Value} \\
\midrule
Learning rate $\alpha$ & 0.001 \\
Mini-batch size & 16 \\
LSTM hidden size & 64 \\
Attention heads & 2 \\
Attention hidden size & 64 \\
FC hidden size & 64 \\
Optimizer & Adam \\
\bottomrule
\end{tabular}
\end{table}

\subsection{Lower-Level Re-Planning}
\label{sec:replan}

When an upper-level candidate solution $\tilde{\mathbf{x}_u}_i$ alters the connectivity structure from the incumbent solution $\mathbf{x}_u^\star$, certain cable routing paths must be recomputed to satisfy the new connectivity constraints. Specifically, connections affected by changes in the upper-level topology—where source or sink substations differ—require replanning; unaffected connections reuse their incumbent lower-level paths $\mathbf{x}_l^\star$ to preserve optimality and computational efficiency. 

Algorithm~\ref{alg:L-Replan} outlines the lower-level replanning procedure, which executes in two phases: (i) identification of affected connections by comparing the new upper-level topology with the incumbent solution, and (ii) sequential re-computation of affected paths via $\mathbb{A}^*$ algorithm. This two-phase structure ensures that unaffected paths benefit from established trench infrastructure while changed paths are optimized under the new connectivity context.

\begin{algorithm}[htb]
\caption{Lower-Level Re-Planning (L-Replan)}
\label{alg:L-Replan}
\begin{algorithmic}[1]
\Require Candidate neighborhood solutions $\{\tilde{\mathbf{x}_u}_1, \ldots, \tilde{\mathbf{x}_u}_N\}$, 
         road graph $G_r$, reference solution $\mathbf{x}_l^\star$
\Ensure Replanned solutions $\{\tilde{\mathbf{x}_l}_1, \ldots, \tilde{\mathbf{x}_l}_N\}$, costs $\{\tilde{F}^{(1)}, \ldots, \tilde{F}^{(N)}\}$

\For{$i = 1$ to $N$}
  \State $\tilde{\mathbf{x}_l}_i \leftarrow \{\}$
  \For{each unchanged connection in $\tilde{\mathbf{x}_u}_i$}
    \State $\tilde{\mathbf{x}_l}_i \leftarrow \tilde{\mathbf{x}_l}_i \cup \mathbf{x}_l^\star(\text{connection})$
  \EndFor
  \For{each changed connection in $\tilde{\mathbf{x}_u}_i$}
    \State $\text{path} \leftarrow \mathbb{A}^*(\text{endpoints}, G_r, \tilde{\mathbf{x}_l}_i)$ 
    \State $\tilde{\mathbf{x}_l}_i \leftarrow \tilde{\mathbf{x}_l}_i \cup \text{path}$
  \EndFor
  \State $\tilde{F}^{(i)} \leftarrow F(\tilde{\mathbf{x}_u}_i, \tilde{\mathbf{x}_l}_i)$
\EndFor

\State \textbf{return} $\{\tilde{\mathbf{x}_l}_1, \ldots, \tilde{\mathbf{x}_l}_N\}, \{\tilde{F}^{(1)}, \ldots, \tilde{F}^{(N)}\}$
\end{algorithmic}
\end{algorithm}

The core mechanism for path replanning is $\mathbb{A}^*$ algorithm that incorporates \textit{trench-sharing awareness}. Unlike classical A$^*$, which evaluates each edge independently, the modified variant dynamically adjusts edge costs based on the incremental accumulation of cable routes within the replanned solution $\tilde{\mathbf{x}_l}_i$. When replanning an affected connection $(s, f)$, the algorithm considers: if edge $(n, m)$ is already traversed by one or more cables in the partially constructed lower-level solution $\tilde{\mathbf{x}_l}_i$, then only the incremental cable installation cost is incurred; otherwise, both trench excavation and cable installation are required.

The total path cost accumulates as the cumulative sum of edge costs along the search trajectory:
\begin{equation}
g(m) = g(n) + \text{cost}(n, m),
\end{equation}
and the evaluation function combines actual accumulated cost with the Manhattan distance heuristic $h(m, t)$:
\begin{equation}
f(m) = g(m) + h(m, t).
\end{equation}

This cost structure automatically incentivizes the algorithm to prefer edges that share existing trenches, thereby capturing the economies of scale inherent in joint trenching.

Notably, the lower-level feasible set $\mathcal{Y}(\mathcal{A}(\mathbf{x}))$ is non-empty for any feasible upper-level solution, as the connected road network always provides at least one routing path between any two required substations. Given the finiteness of the problem domain, lower-level optimization always admits an optimal solution. Feasibility is ensured by construction: only affected connections are recomputed via modified $\mathbb{A}^*$ with dynamic capacity checking, while unaffected connections reuse incumbent paths. Upper-level candidate solutions violating edge capacity constraints are directly discarded without invoking lower-level optimization.

\subsection{Solution Evaluation and Update}
\label{sec:update}

Upon completing the lower-level re-planning phase, the best candidate solution is identified as:
\begin{equation}
(\hat{\mathbf{x}_u}, \hat{\mathbf{x}_l}, \hat{F}) \leftarrow \arg\min_{i=1}^{N} \tilde{F}^{(i)}.
\end{equation}
If $\hat{F} < F^\star$, the incumbent solution is updated accordingly. The operator weight vector $w = [w_1, w_2, w_3]$ is then adaptively adjusted via:
\begin{equation}
w_j \leftarrow w_j \cdot \begin{cases}
\rho_{\text{succ}}, & \text{if operator } j \text{ improves the solution}, \\
\rho_{\text{fail}}, & \text{otherwise},
\end{cases}
\end{equation}
where $\rho_{\text{succ}} = 1.1$ and $\rho_{\text{fail}} = 0.98$ are the success and failure decay factors. Every 20 iterations, $w$ is re-initialized to $[1, 1, 1]$ to restore exploration balance. This adaptive scheme ensures high-performing operators receive increased selection probability in subsequent iterations, thereby balancing search exploitation and exploration.

\subsection{Algorithmic Complexity Analysis of L-VNS}
\label{sec:complexity}

Let $N_s$ denote the number of substations, $F$ the number of feeders, $V$ and $E$ the vertices and edges in the road graph, $P$ and $G$ the HGS population size and generations, $T$ the VNS iterations, and $N$ the candidate samples per iteration. We characterize the main components as follows.

The HGS-based initialization cost is dominated by local-search applications across the population and generations: $O(P\cdot G\cdot L_{\mathrm{loc}})$, where $L_{\mathrm{loc}}=O(N_s^2)$ for neighborhood evaluations, yielding $O(PGN_s^2)$ overall. Each candidate evaluation in the L-VNS main loop requires computing shortest paths for $O(N_s)$ point-to-point routes on the road graph; a single $\mathbb{A}^*$ invocation costs $O(E\log V)$, so the per-evaluation routing cost is $O(N_sE\log V)$. The L-VNS loop performs $T$ iterations, each evaluating $N$ candidates with refinement cost $L_{\mathrm{loc}}$ and routing cost $N_sE\log V$, giving a total cost of $O \big( TN({N_s}^2+ N_s E\log V)\big)$. The DRL policy inference cost is negligible relative to routing and local search.

The total worst-case cost is $O\big( PG{N_s}^2+TN({N_s}^2+N_s E \log V) \big)$, where the term $TNN_sE \log V$ dominates. Moreover, routing evaluations are highly parallelizable, and $\mathbb{A}^*$ typically performs much better than worst-case on sparse urban road graphs.

\section{Experimental Settings}
\label{sec:settings}
Experiments were conducted on a PC equipped with an AMD Ryzen 7 5800H CPU (3.2 GHz) and an NVIDIA GeForce RTX 3060 GPU. All algorithms were implemented in Python 3.10. The source code will be released upon publication of the paper. This section details the benchmark instances, competitive baselines, and ablated variants used for evaluation and comparison.

\subsection{Benchmark Construction}
\label{sec:benchmark}
Following literature \cite{bayliss2012transmission, gu2019full}, we define the key parameters for our simulations. The peak demand \(q_i\) at each MV substation \(i\) is independently and uniformly drawn from the range \([2, 5]\) MVA. Each feeder is assigned a rated capacity of \(Q = 10\) MVA. The maximum allowable number of parallel cables on any given edge is \(C_{e}^{\max} = 6\). Unit costs are specified as 1.5 million CNY/km for road trenching \((c_{e}^{\mathrm{tr}})\) and 0.5 million CNY/km for cable laying on edge \(e\) \((c_{e}^{\mathrm{ca}})\).

To thoroughly evaluate the proposed algorithms in a controlled yet scalable environment, we construct a suite of benchmark instances based on regular square lattices. The size of the grid dictates the scale of the instance, while its orthogonal geometry effectively approximates typical block street layouts. Specifically, an \(n_{grid}\times n_{grid}\) lattice defines \(n_{grid}^{2}\) uniformly arranged blocks, each spanning \(1\ \text{km} \times 1\ \text{km}\). The locations of HV and MV substations are determined through a two-level clustering process. Initially, each lattice node is considered a potential load aggregation point. A K-means algorithm then partitions all these nodes into \(N_{\mathrm{mv}}\) clusters, with the centroids of these clusters serving as candidate MV substation locations. Subsequently, these MV candidate locations are reclustered into \(N_{\mathrm{hv}}\) groups, from which the HV substation locations are derived. As cluster centroids may not precisely align with lattice vertices, each centroid is projected onto the closest point on the nearest lattice edge to define the final substation coordinates.

This study investigates four problem instances defined by the tuple $(n_{grid}, N_{\text{MV}}, N_{\text{HV}}) \in $ \{(20, 30, 5), (30, 50, 6), (30, 80, 7), (30, 100, 8)\}, representing typical 10\,kV urban distribution networks. The grid sizes of $20 \times 20$\,km and $30 \times 30$\,km cover urban areas from small city centers to large metropolitan regions, while the MV/HV substation counts (30--100 and 5--8, respectively) reflect realistic network densities. For each scale, three independent instances were generated with random seed 42, yielding a benchmark suite of 12 instances labeled as $\text{Case}\,i\text{-}j$, where $i \in \{1,2,3,4\}$ denotes the scale and $j \in \{1,2,3\}$ denotes the instance replicate. The corresponding data and generation code are available at \footnote{https://github.com/Will-iam-L/L-MVNS-for-urban-cable-routing}.

\subsection{Competitive Baselines}
\label{sec:Baselines}
For comparative analysis, several representative baselines are implemented:(1) The modified two-phase Clark-Wright savings algorithm (\textbf{MCWS}) \cite{wang2024practical}, proposed in 2024, serves as a relation-only baseline. It optimizes substation connectivity by utilizing precomputed shortest-path distances, explicitly disregarding trench sharing and thus equating trench length with cable length; (2) The hybrid genetic search algorithm (\textbf{HGS}) \cite{VIDAL2022105643}, which effectively combines a genetic algorithm with local search techniques. This algorithm is known for its strong performance on COPs, such as vehicle routing problems; (3) General Variable Neighborhood Search (\textbf{GVNS}): An ablated variant of L-VNS that employs uniform random sampling during neighborhood construction instead of DRL-guided adaptive sampling; and surrogate-assisted neighborhood search (SANS) algorithms, including (4) \textbf{SANS-Kriging} \cite{8477763} and (5) \textbf{SANS-MLP} \cite{ren2022ensemble}. \textbf{SANS-Kriging} leverages a Gaussian process surrogate to directly approximate the lower-level optimal cost function, enabling rapid bilevel evaluation at the expense of reduced approximation fidelity. \textbf{SANS-MLP} extends this paradigm by replacing the statistical surrogate with a multi-layer perceptron trained via supervised learning on offline bilevel solutions, achieving higher expressiveness in capturing complex topology-routing dependencies, though incurring additional training overhead and potential over-fitting risks in data-sparse regimes.

Regarding algorithmic adaptation and parameter specification: MCWS follows settings from its original work \cite{wang2024practical}; HGS utilizes the PYVRP library \cite{Wouda_Lan_Kool_PyVRP_2024} with factory defaults. As surrogate-assisted nested search has not yet been applied to urban cable routing, we propose a unified two-level solution framework. At the upper level, we employ the U-VNS procedure (identical to L-VNS to isolate the learning component's contribution). At the lower level, pre-trained surrogate models (Kriging or MLP) are utilized to predict objective function values based on upper-level solutions, thereby circumventing explicit lower-level optimization. The surrogate inputs are represented as upper-level decisions encoded as one-dimensional node-index tensors. The surrogates are initially trained during the first upper-level iteration, where a high-fidelity heuristic (employing a neighborhood size of 10, with destructive operators $\mathbb{O}_1$ randomly applied at 1-5 selected positions, followed by lower-level re-planning, executed for 20 iterations in total) serves as the lower-level optimizer to generate training labels (i.e., objective function values). Subsequently, the surrogates are incrementally updated with new high-fidelity evaluations as they become available throughout the iterative search process.

For SANS-Kriging: We employ Kriging with constant and RBF kernels, tuning amplitude and length scale within $[10^{-3}, 10^3]$ via 10 restarts. The Gaussian process regressor is trained offline using maximum likelihood estimation on a set of reference bilevel solutions, and then incrementally updated with new high-fidelity evaluations. For SANS-MLP: A 3-layer MLP (32 hidden neurons, ReLU activations, 0.2 dropout) is trained with the Adam optimizer ($\alpha = 0.001$) and mean squared error (MSE) loss. The training employs mini-batch gradient descent with batch size 32 and shuffled sampling over 10 epochs. For iterative algorithms (SANS-Kriging, SANS-MLP, and L-VNS), initialization times were set to 100, 200, 300, and 400 seconds, respectively, while maximum search times were restricted to 500, 1000, 1500, and 2000 seconds, respectively. HGS, employing random initialization, is allocated a total running time equivalent to the sum of the maximum search times provided to these iterative algorithms for fair comparison. Each algorithm was evaluated using 10 independent runs per test case.

\subsection{Ablated Variants}
\label{sec:Ablated Variants}

To systematically evaluate the contribution of individual components within the L-VNS framework, we introduce a series of ablated variants based on the full L-VNS method:

\begin{enumerate}
    \item \textbf{AM-1 (No Auxiliary Initialization):} uses random initialization, bypassing the HGS-based auxiliary task construction. This variant isolates the impact of high-quality solution initialization on overall algorithm performance.
    
    \item \textbf{AM-2-1, AM-2-2, AM-2-3 (Two-Operator Variants):} Each variant retains only two neighborhood operators while removing the other one. Specifically, AM-2-1 removes $\mathbb{O}_1$ (path-destruction), AM-2-2 removes $\mathbb{O}_2$ (intra-feeder 2-opt), and AM-2-3 removes $\mathbb{O}_3$ (inter-feeder 2-opt). These variants elucidate the individual contribution of each destruction operator to the overall solution quality.
    
    \item \textbf{AM-3 (No DRL Guidance):} the same as the aforementioned GVNS, uses uniform random neighborhood sampling, replacing the DRL-guided probability-based sampling mechanism. This variant quantifies the benefit of integrating the multi-agent DRL module for intelligent neighborhood generation.
    
    \item \textbf{AM-4 (Classical $A^*$ Repair):} uses classical A$^*$ algorithm for lower-level path replanning, replacing the modified $\mathbb{A}^*$ that incorporates trench-sharing awareness. This variant demonstrates the importance of accounting for shared trenching costs during path optimization.
\end{enumerate}

\section{Experiments and Results}
\label{results}

\subsection{Effectiveness Comparison}
\label{Effectiveness Comparison}

\begin{figure*}[H]
	\centering
	\includegraphics[width=1\linewidth]{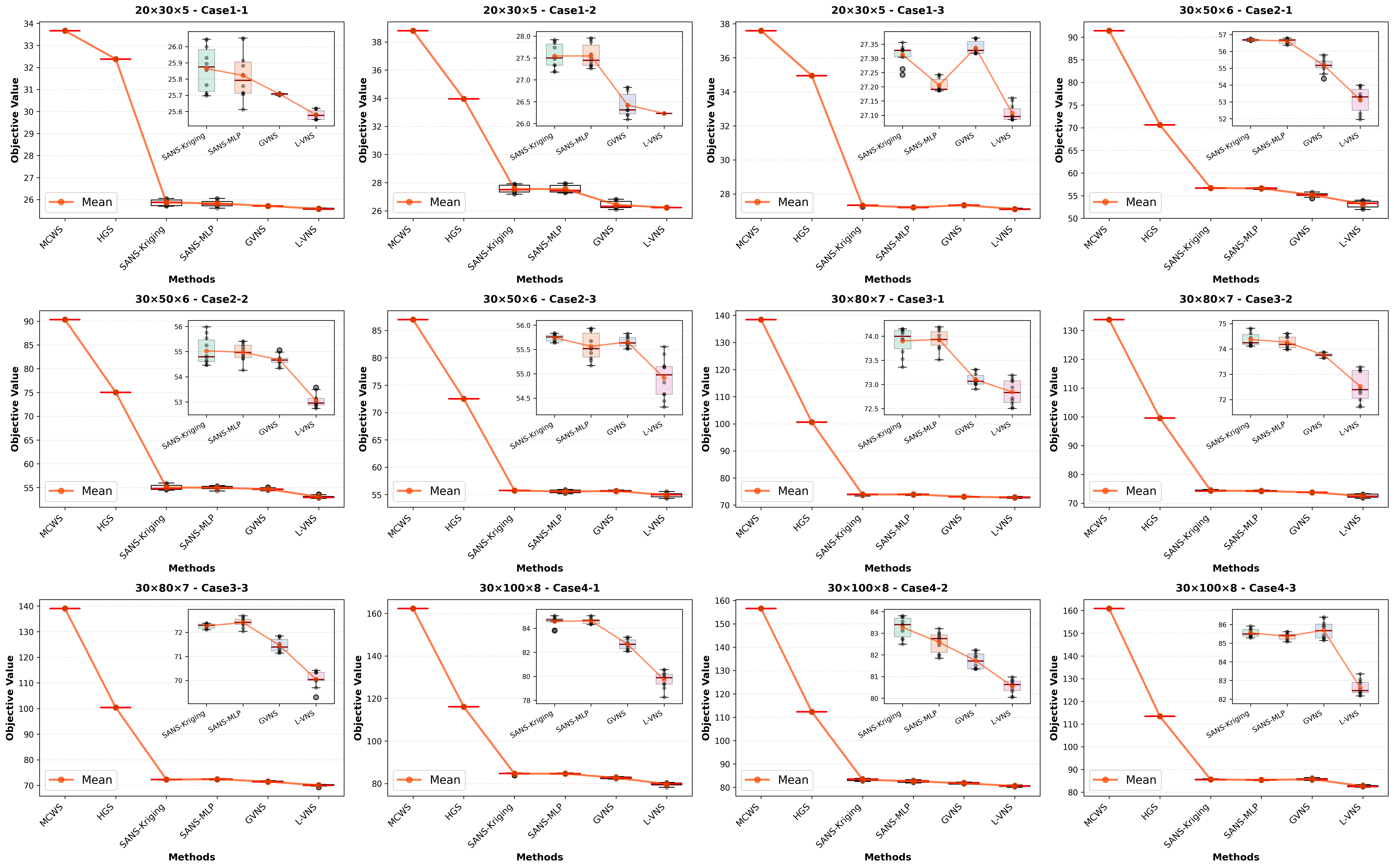}
    \caption{Boxplot distributions of solution costs (million CNY) across 12 test instances over 10 independent runs. Boxes represent interquartile ranges with median lines; whiskers indicate data extremes; red dots mark means. An inset subplot (upper right) magnifies the four best-performing methods for clearer performance discrimination.}
	\label{fig: COMPARE_ALL}
\end{figure*}

Table~\ref{table: result_all} reports the mean solution costs (million CNY) and corresponding standard deviations (in parentheses) achieved by each algorithm across 10 independent runs on all test instances. The best performance on each instance is highlighted in bold. Statistical significance is indicated using standard notation: $+$ (baseline significantly better), $-$ (L-VNS significantly better), and $\approx$ (no significant difference). Wilcoxon signed-rank test with $p < 0.05$ is used for the significance test. The final rows present aggregate significance counts $(+/\approx/-)$ and average relative percentage deviation (RPD) relative to L-VNS. Fig.~\ref{fig: COMPARE_ALL} provides complementary boxplot visualizations of the cost distributions, revealing algorithm robustness and variance profiles.

\begin{table*}[ht]
\centering
\setlength{\tabcolsep}{3.5pt}
\caption{Objective-value comparison between L-VNS and baseline algorithms. Reported values are mean construction costs (million CNY) over 10 independent runs, with standard deviations in parentheses. Bold indicates the best result.}
\begin{tabular}{cc|c|c|c|c|c|c}
\hline
\multicolumn{2}{c|}{\textbf{Problem}}                     & \textbf{MCWS}     & \textbf{HGS}      & \textbf{SANS-Kriging} & \textbf{SANS-MLP} & \textbf{GVNS}     & \textbf{L-VNS}         \\
\textbf{Scale}                       & \textbf{Instance}  & \textbf{}         & \textbf{}         & \textbf{}             & \textbf{}         & \textbf{}         & \textbf{(Ref.)}        \\ \hline
\multirow{3}{*}{\textbf{20×30×5}}    & Case1-1            & 33.67   (0.00)-   & 32.38   (0.00)-   & 25.86   (0.10)-      & 25.82   (0.10)-  & 25.71   (0.00)-   & \textbf{25.58} (0.00)  \\
                                     & Case1-2            & 38.80   (0.00)-   & 33.96   (0.00)-   & 27.54   (0.30)-      & 27.55   (0.30)-  & 26.43   (0.30)$\approx$   & \textbf{26.23} (0.00)  \\
                                     & Case1-3            & 37.59   (0.00)-   & 34.95   (0.00)-   & 27.31   (0.00)-       & 27.21   (0.00)-   & 27.34   (0.00)-   & \textbf{27.11} (0.00)  \\ \hline
\multirow{3}{*}{\textbf{30×50×6}}    & Case2-1            & 91.42   (0.00)-   & 70.65   (0.00)-   & 56.68   (0.10)-       & 56.62   (0.10)-  & 55.17   (0.40)-   & \textbf{53.13} (0.80) \\
                                     & Case2-2            & 90.32   (0.00)-   & 75.04   (0.00)-   & 55.03   (0.60)-       & 54.98   (0.40)-  & 54.67   (0.20)-   & \textbf{53.05} (0.30) \\
                                     & Case2-3            & 86.99   (0.00)-   & 72.51   (0.00)-   & 55.74   (0.10)-       & 55.56   (0.30)-  & 55.65   (0.10)-   & \textbf{54.91} (0.40) \\ \hline
\multirow{3}{*}{\textbf{30×80×7}}    & Case3-1            & 138.44   (0.00)-  & 100.61   (0.00)-  & 73.90   (0.30)-      & 73.93   (0.20)-  & 73.10   (0.10)-   & \textbf{72.84} (0.30) \\
                                     & Case3-2            & 133.76   (0.00)-  & 99.56   (0.00)-   & 74.37   (0.30)-      & 74.26   (0.30)-  & 73.76   (0.10)-   & \textbf{72.52} (0.60) \\
                                     & Case3-3            & 139.04   (0.00)-  & 100.42   (0.00)-  & 72.27   (0.10)-       & 72.41   (0.20)-  & 71.46   (0.30)-   & \textbf{70.05} (0.30) \\ \hline
\multirow{3}{*}{\textbf{30×100×8}}   & Case4-1            & 162.31   (0.00)-  & 116.10   (0.00)-  & 84.57   (0.40)-      & 84.63   (0.30)-  & 82.67   (0.40)-   & \textbf{79.72} (0.70) \\
                                     & Case4-2            & 156.56   (0.00)-  & 112.31   (0.00)-  & 83.27   (0.50)-      & 82.58   (0.50)-  & 81.74   (0.40)-   & \textbf{80.55} (0.30) \\
                                     & Case4-3            & 160.90   (0.00)-  & 113.39   (0.00)-  & 85.55   (0.20)-      & 85.37   (0.20)-  & 85.70   (0.50)-   & \textbf{82.62} (0.40) \\ \hline
\multicolumn{2}{c|}{\textbf{Significance ($+/\approx/-$)}} & \textbf{(0/0/12)} & \textbf{(0/0/12)} & \textbf{(0/0/12)}     & \textbf{(0/0/12)} & \textbf{(0/1/11)} & ---             \\ \hline
\multicolumn{2}{c|}{\textbf{Average RPD}}                 & \textbf{73.72\%}  & \textbf{36.04\%}  & \textbf{3.24\%}       & \textbf{3.08\%}   & \textbf{1.95\%}   & ---       \\ \hline
\end{tabular}
\label{table: result_all}
\end{table*}

\textbf{Path-Aware vs. Relation-Only Methods.} Path-aware baselines (SANS-Kriging, SANS-MLP, GVNS, and L-VNS) greatly reduce costs relative to relation-only baselines (MCWS and HGS). Specifically, MCWS exhibits an average RPD of 73.72\%, while HGS achieves 36.04\%. This substantial gap arises because road-trenching costs typically exceed cable material costs by approximately threefold in engineering practice \cite{deveci2019electrical}, and relation-only methods fail to capture trench-sharing opportunities by ignoring explicit cable routing paths. MCWS further suffers from its multi-stage decomposition strategy, where early greedy decisions (e.g., sweep-based clustering and Clarke–Wright sequencing) preclude globally optimal connectivity structures.

\textbf{Surrogate-Assisted Methods.} SANS-Kriging and SANS-MLP achieve average RPDs of 3.24\% and 3.08\%, respectively, indicating moderate competitiveness. These methods accelerate bilevel evaluation by approximating the lower-level optimal cost function using Kriging or MLP surrogates. Although widely adopted for bilevel optimization in continuous domains, surrogate-assisted approaches face inherent limitations when applied to large-scale combinatorial problems. First, the exponentially large discrete state space makes it difficult to guarantee approximation fidelity—even minor upper-level topology changes can trigger substantial lower-level routing reconfiguration, challenging surrogates trained on finite samples. Second, surrogate performance critically depends on the coverage of samples collected during training; insufficient exploration of the feasible region leads to poor extrapolation when encountering novel solution structures. Third, incremental surrogate updates during search incur additional computational overhead for retraining, partially offsetting the speedup from avoiding exact lower-level optimization. These limitations collectively prevent surrogate methods from fully capturing the dynamic trench-sharing dependencies, representing critical challenges for future research.

\textbf{GVNS vs. L-VNS.} GVNS, which employs uniform random neighborhood sampling, achieves an average RPD of 1.95\% and is lower than other competitor baselines. This competitive performance validates the effectiveness of the proposed framework without the DRL module. However, L-VNS consistently outperforms GVNS across all test cases, demonstrating the added value of DRL-guided neighborhood sampling. The advantage of L-VNS becomes more pronounced on larger instances, where the combinatorial search space expands exponentially. This trend suggests that DRL agents effectively learn to prioritize high-quality perturbation positions by encoding both upper-level connectivity patterns and lower-level routing structures, thereby reducing futile bilevel evaluations and accelerating convergence.

\textbf{Scalability Trends.} Although the DRL module in L-VNS is trained exclusively on small-scale instances with configuration $(20, 30, 5)$, it significantly outperforms the non-learning baseline GVNS and other competing methods on unseen instances with problem scales ranging from $(20, 30, 5)$ to $(30, 100, 8)$. This scalability demonstrates that the DRL agents learn generalizable behavioral policies for this class of problems, rather than instance-specific solutions tailored to particular problem configurations. Moreover, such generalization capability is further enhanced by the transfer learning mechanism applied to the test instances. Across the four problem scales examined in this study, L-VNS exhibits robust generalization performance without significant degradation as problem scale increases.

In summary, L-VNS achieves the best overall performance across all 12 benchmark instances. The integration of auxiliary task-based initialization, variable neighborhood search with DRL-guided sampling, and incremental lower-level re-planning enables L-VNS to efficiently navigate the exponentially large bilevel search space while maintaining superior solution quality and algorithmic stability.

\subsection{Ablation Experiments}
\label{Ablation Experiments}

To systematically evaluate the contribution of each algorithmic component and justify design choices in the proposed L-VNS framework, we conducted comprehensive ablation studies on all 12 benchmark instances. Following the introduction presented in Section~\ref{sec:Ablated Variants}, we implemented six variant configurations: AM-1, AM-2-1, AM-2-2, AM-2-3, AM-3, and AM-4. Table~\ref{tab:ablation-results} presents comprehensive performance comparisons across all variants, reporting mean objective function values, standard deviations (in parentheses), Wilcoxon signed-rank test significance indicators ($+/\approx/-$), and average RPD relative to L-VNS. These results quantitatively demonstrate the cumulative impact of each component and validate the necessity of integrating learning-guided neighborhood exploration within the variable neighborhood search framework.

\begin{table*}[ht]
\centering
\setlength{\tabcolsep}{1.5pt}
\caption{Ablation study comparing L-VNS against its variants. Entries denote mean objective costs (million CNY) and standard deviations (in parentheses) across 10 independent runs. Bold entries highlight best performance.}
\label{tab:ablation-results}
\begin{tabular}{cc|c|c|c|c|c|c|c}
\hline
\multicolumn{2}{c|}{\textbf{Problem}}                     & \textbf{AM-1}     & \textbf{AM-2-1}   & \textbf{AM-2-2}   & \textbf{AM-2-3}   & \textbf{AM-3}     & \textbf{AM-4}     & \textbf{L-VNS}         \\
\textbf{Scale}                       & \textbf{Instance}  & \textbf{}         & \textbf{}         & \textbf{}         & \textbf{}         & \textbf{}         & \textbf{}         & \textbf{(Ref.)}        \\ \hline
\multirow{3}{*}{\textbf{20×30×5}}    & Case1-1            & 47.87   (0.20)-   & 26.03   (0.20)-   & 25.89   (0.00)-   & 26.00   (0.10)-   & 25.71   (0.00)-   & 26.83   (0.30)-   & \textbf{25.58} (0.00)  \\
                                     & Case1-2            & 56.85   (0.20)-   & 26.44   (0.30)$\approx$   & 26.35   (0.20)$\approx$   & 28.01   (0.20)-   & 26.43   (0.30)$\approx$   & 27.20   (0.30)-   & \textbf{26.23} (0.00)  \\
                                     & Case1-3            & 46.25   (0.50)-   & 27.41   (0.00)-   & 27.32   (0.00)-   & 27.24   (0.10)-   & 27.34   (0.00)-   & 27.23   (0.20)$\approx$   & \textbf{27.11} (0.00)  \\ \hline
\multirow{3}{*}{\textbf{30×50×6}}    & Case2-1            & 99.55   (0.30)-   & 55.64   (0.20)-   & 55.43   (0.40)-   & 56.70   (0.10)-   & 55.17   (0.40)-   & 58.07   (0.20)-   & \textbf{53.13} (0.80) \\
                                     & Case2-2            & 105.66   (0.50)-  & 56.43   (0.30)-   & 54.79   (0.10)-   & 54.60   (0.30)-   & 54.67   (0.20)-   & 56.09   (0.50)-   & \textbf{53.05} (0.30) \\
                                     & Case2-3            & 106.10   (0.60)-  & 56.97   (0.40)-   & 56.07   (0.30)-   & 55.68   (0.40)-   & 55.65   (0.10)-   & 56.84   (0.10)-   & \textbf{54.91} (0.40) \\ \hline
\multirow{3}{*}{\textbf{30×80×7}}    & Case3-1            & 135.24   (0.60)-  & 74.66   (0.30)-   & 72.97   (0.10)$\approx$   & 73.17   (0.30)$\approx$   & 73.10   (0.10)-   & 74.83   (0.30)-   & \textbf{72.84} (0.30) \\
                                     & Case3-2            & 148.18   (0.30)-  & 74.59   (0.10)-   & 73.77   (0.20)-   & 74.09   (0.30)-   & 73.76   (0.10)-   & 75.05   (0.10)-   & \textbf{72.52} (0.60) \\
                                     & Case3-3            & 132.73   (0.70)-  & 73.65   (0.30)-   & 71.81   (0.20)-   & 71.98   (0.20)-   & 71.46   (0.30)-   & 73.47   (0.30)-   & \textbf{70.05} (0.30) \\ \hline
\multirow{3}{*}{\textbf{30×100×8}}   & Case4-1            & 163.25   (1.00)-  & 83.63   (0.30)-   & 82.37   (0.20)-   & 84.45   (0.10)-   & 82.67   (0.40)-   & 84.52   (0.20)-   & \textbf{79.72} (0.70) \\
                                     & Case4-2            & 174.80   (0.80)-  & 83.21   (0.20)-   & 81.85   (0.20)-   & 82.70   (0.10)-   & 81.74   (0.40)-   & 83.53   (0.30)-   & \textbf{80.55} (0.30) \\
                                     & Case4-3            & 163.21   (0.30)-  & 84.84   (0.20)-   & 85.72   (0.40)-   & 85.93   (0.60)-   & 85.70   (0.50)-   & 87.06   (0.10)-   & \textbf{82.62} (0.40) \\ \hline
\multicolumn{2}{c|}{\textbf{Significance ($+/\approx/-$)}} & \textbf{(0/0/12)} & \textbf{(0/1/11)} & \textbf{(0/2/10)} & \textbf{(0/1/11)} & \textbf{(0/1/11)} & \textbf{(0/1/11)} & ---            \\ \hline
\multicolumn{2}{c|}{\textbf{Average RPD}}                 & \textbf{96.09\%}  & \textbf{3.32\%}   & \textbf{2.11\%}   & \textbf{3.16\%}   & \textbf{1.95\%}   & \textbf{4.48\%}   & ---        \\ \hline
\end{tabular}
\end{table*}

\textbf{Impact of Auxiliary Task Initialization (AM-1 vs. L-VNS).} AM-1, which employs random initialization instead of the HGS-based auxiliary task, exhibits an average RPD of 96.09\% and significantly lags behind on all 12 instances. This dramatic degradation underscores the critical importance of high-quality initial solutions in bilevel optimization. Random initialization forces subsequent search to simultaneously address extensive upper-level exploration and non-negligible lower-level re-planning from an inferior starting point, severely compromising search efficiency. In contrast, L-VNS with auxiliary task-assisted initialization efficiently generates high-quality feasible solutions by focusing solely on upper-level connectivity without invoking lower-level routing optimization, thereby providing superior starting points that substantially accelerate neighborhood search convergence.

\textbf{Contribution of Individual Neighborhood Operators (AM-2-1, AM-2-2, AM-2-3 vs. L-VNS).} The two-operator variants achieve average RPDs of 3.32\%, 2.11\%, and 3.16\%, with each being significantly worse than L-VNS on 11 of 12 instances, demonstrating that the three operators address complementary aspects: $\mathbb{O}_1$ enables large-scale connectivity reconfiguration; $\mathbb{O}_2$ optimizes edge-crossing patterns within feeders; and $\mathbb{O}_3$ balances load distribution and discovers inter-feeder trench-sharing opportunities. Their synergistic integration allows L-VNS to escape local optima more effectively than any single mechanism alone.

\textbf{Value of DRL-Guided Sampling (AM-3 vs. L-VNS).} AM-3, which replaces DRL-guided sampling with uniform random selection, attains an average RPD of 1.95\%. As noted earlier, this highlights that effective sampling is crucial to avoid wasted evaluations: uniform sampling explores all regions equally regardless of past performance, while DRL focuses on high-potential neighborhoods.

\textbf{Importance of Trench-Sharing-Aware Routing (AM-4 vs. L-VNS).} AM-4, which uses classical A$^*$ without dynamic edge-cost adjustment for trench sharing, achieves an average RPD of 4.48\% and is worse on all 12 instances. Classical A$^*$ relies on static edge costs and cannot capture trench-sharing opportunities, which are critical since trenching costs typically exceed cable material costs. In contrast, the modified $\mathbb{A}^*$ dynamically adjusts edge costs during sequential cable placement to incentivize parallel deployment and reduce overall trenching expenses.

In summary, L-VNS's superior performance stems from the coordinated integration of four key components: auxiliary task initialization, upper-level VNS, DRL-guided sampling, and lower-level re-planing. Each component addresses a distinct computational or modeling challenge, and their synergy is essential for achieving state-of-the-art solution quality.

\subsection{Sensitivity Analysis}
This subsection examines how two key hyperparameters affect L-VNS performance: initialization time allocated to the auxiliary task and neighborhood size per search operator. These experiments are conducted on the medium-scale test case Case3-1, with 10 independent runs performed for each parameter configuration.

\subsubsection{Initialization Time}

\begin{figure}
	\centering
	\includegraphics[width=1\linewidth]{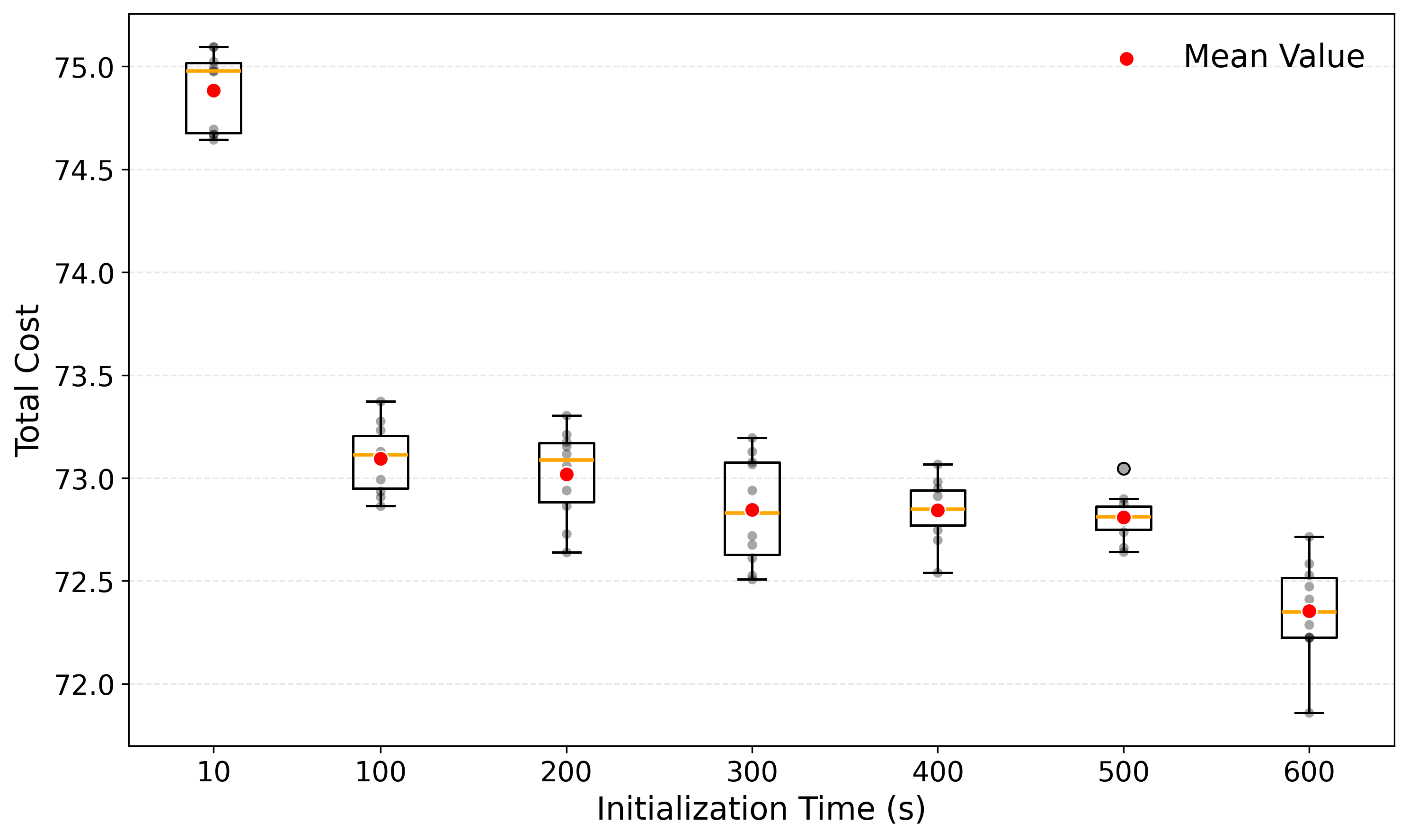}
	\caption{Boxplots of L-VNS performance over 10 runs with initialization time from 10 to 600 seconds.}
	\label{fig: initialization time}
\end{figure}

We varied the initialization time (i.e., auxiliary task optimization) from 10 to 600 seconds to assess its impact on the subsequent search. The maximum iteration count is fixed at 400. Fig. \fref{fig: initialization time} reports the performance on each scene. In general, longer initialization yields faster convergence during the neighborhood search; however, beyond 200 seconds, the marginal reduction in total cost plateaus. The mean performance at each initialization time (i.e., red dots) approximately follows an exponential decay trend. Notably, in fact, a longer initialization does not necessarily produce a better initial solution because the auxiliary task does not explicitly account for road‑trenching costs. Therefore, setting the initialization time to 300 seconds for this case represents a reasonable trade-off between solution quality and computational efficiency.

\subsubsection{Neighborhood Size}
\begin{figure}
	\centering
	\includegraphics[width=1\linewidth]{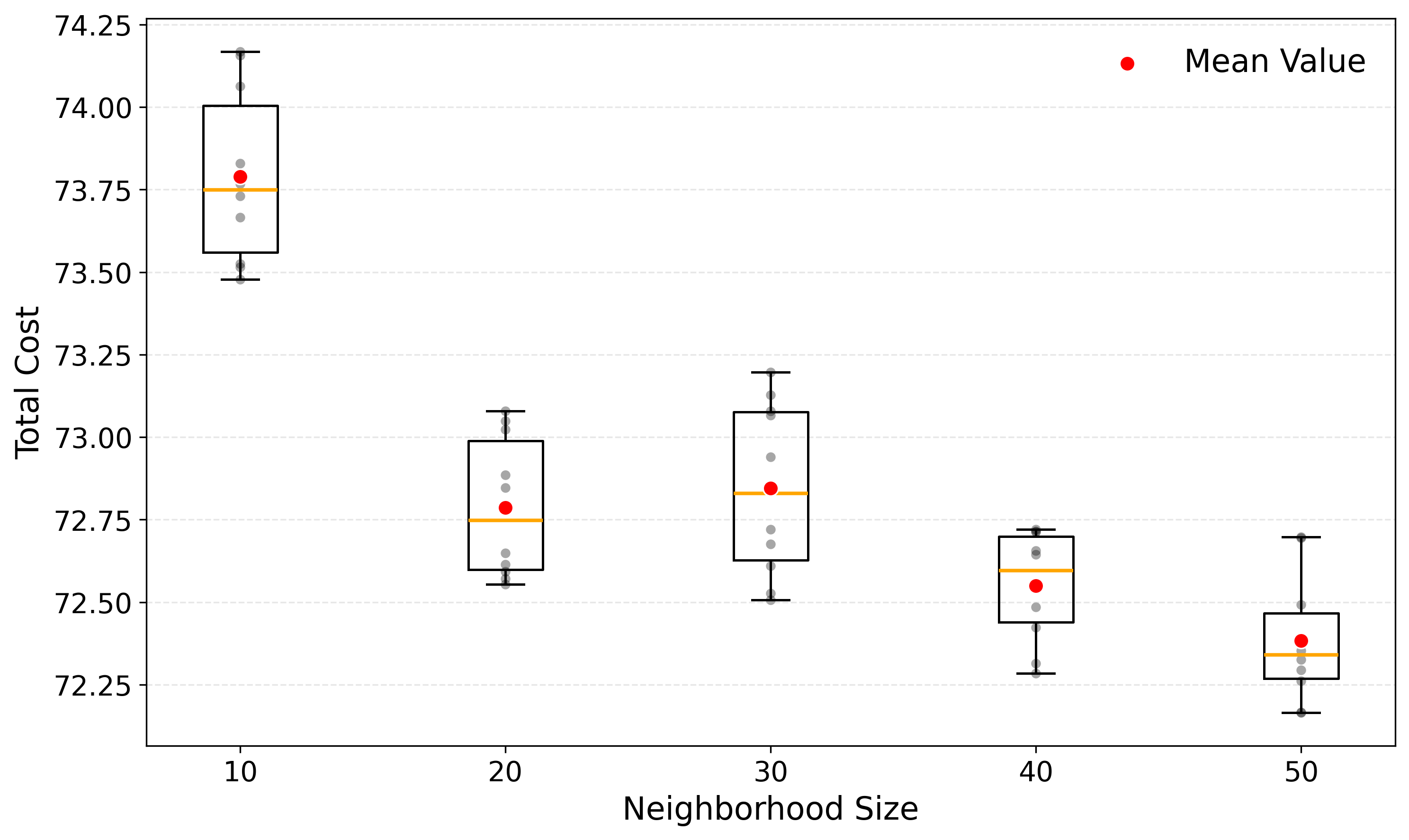}
	\caption{Boxplots of L-VNS performance over 10 runs with neighborhood size per operator from 10 to 50.}
	\label{fig: neighbor size}
\end{figure}

Expanding the neighborhood search range can improve the likelihood of reaching a global optimum, but it also increases computational cost and runtime. In this subsection, we evaluate L-VNS under different neighborhood sizes to identify a suitable trade-off. We vary the number of candidate sizes \(N\) from 10 to 50. Fig. \fref{fig: neighbor size} reports its performance across these settings.

The average cost drops markedly when increasing \(N\) from 10 to 20. Beyond \(N=30\), solution quality plateaus while runtime grows linearly with \(N\). Moreover, very large neighborhoods can dilute the benefits of DRL guidance because the acceptance pool increasingly includes low‑quality, randomly perturbed candidates. Therefore, balancing accuracy and efficiency, we set the neighborhood size per operator to 30.

\subsection{Case Study on Real Urban Networks}
To assess the practical applicability of L-VNS on realistic urban networks, we use three GIS-derived OpenStreetMap test cases from Changsha, Hunan, China, including: the core Kaifu District (Case5), the core Wangcheng District (Case6), and the core East-Changsha (Case7). These cases feature $(n_{\text{MV}}, n_{\text{HV}}) \in \{(30, 5), (50, 6), (80, 7)\}$ respectively, with computational budgets matching Cases 2 to 4. Similar to the grid-derived test cases introduced in \sref{sec:benchmark}, these GIS-derived cases are generated via hierarchical K-means clustering on OpenStreetMap road networks and building footprints in Changsha, Hunan Province. Building centroids are partitioned into $n_{\text{MV}}$ clusters, then recursively clustered into $n_{\text{HV}}$ groups. Substation locations are snapped to the nearest road network node, and all geometric operations are performed in UTM projected coordinates for metric accuracy. Each case is tested 10 times independently against three competitive baselines: SANS-Kriging, SANS-MLP, and GVNS. The comparison results are shown in Table~\ref{tab:gis-results}.

\begin{table}[ht]
\centering
\footnotesize
\setlength{\tabcolsep}{4pt}
\caption{Comparison results on three GIS-derived OpenStreetMap test cases. Mean construction costs (million CNY) are reported from 10 independent runs, with corresponding standard deviations shown in parentheses.}
\label{tab:gis-results}
\begin{tabular}{c|c|c|c|c}
\hline
\textbf{Instance} & \textbf{SANS-Kriging} & \textbf{SANS-MLP} & \textbf{GVNS} & \textbf{L-VNS} \\
\hline
\textbf{Case5} & \makecell{61.73 \\ (0.30)-} & \makecell{61.10 \\ (0.10)-} & \makecell{60.52 \\ (0.00)-} & \makecell{\textbf{59.88} \\ (0.20)} \\

\textbf{Case6} & \makecell{117.94 \\ (0.60)-} & \makecell{117.81 \\ (0.70)-} & \makecell{118.56 \\ (0.70)-} & \makecell{\textbf{116.96} \\ (0.70)} \\

\textbf{Case7} & \makecell{165.18 \\ (0.20)-} & \makecell{164.56 \\ (0.40)-} & \makecell{163.95 \\ (0.10)-} & \makecell{\textbf{163.40} \\ (0.20)} \\
\hline
\makecell{\textbf{Significance} \\ ($+/\approx/-$)} & \textbf{(0/0/3)} & \textbf{(0/0/3)} & \textbf{(0/0/3)} & --- \\
\hline
\makecell{\textbf{Average} \\ \textbf{RPD}} & \textbf{1.67\%} & \textbf{1.16\%} & \textbf{0.92\%} & --- \\
\hline
\end{tabular}
\end{table}

\begin{figure*}[!htbp]
  \centering
  
  \begin{minipage}{0.31\linewidth}
    \centering
    \includegraphics[width=\linewidth]{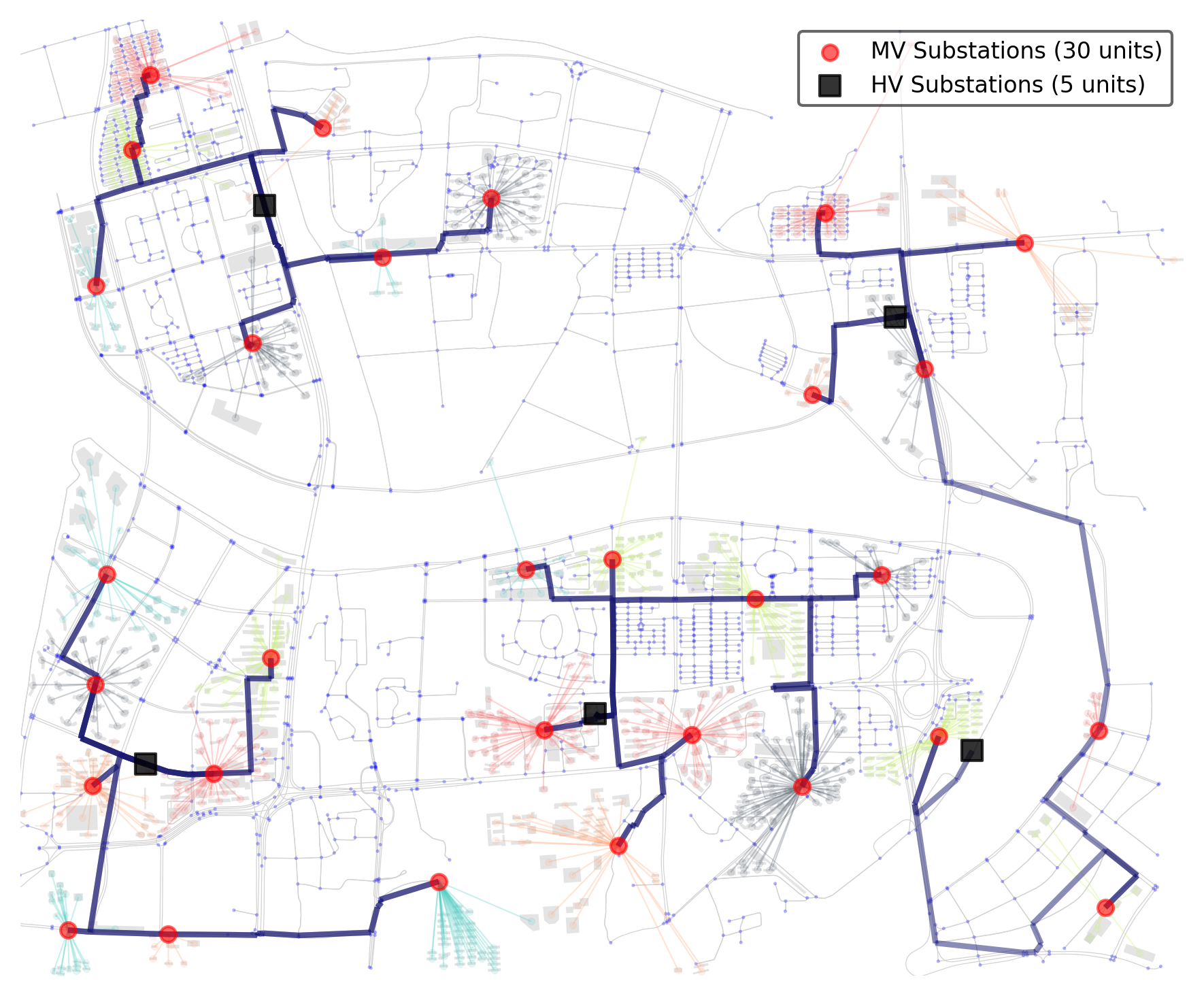}
    \\ \textbf{(a)} Solution for Case5 (Kaifu)
  \end{minipage}
  \hfill
  \begin{minipage}{0.35\linewidth}
    \centering
    \includegraphics[width=\linewidth]{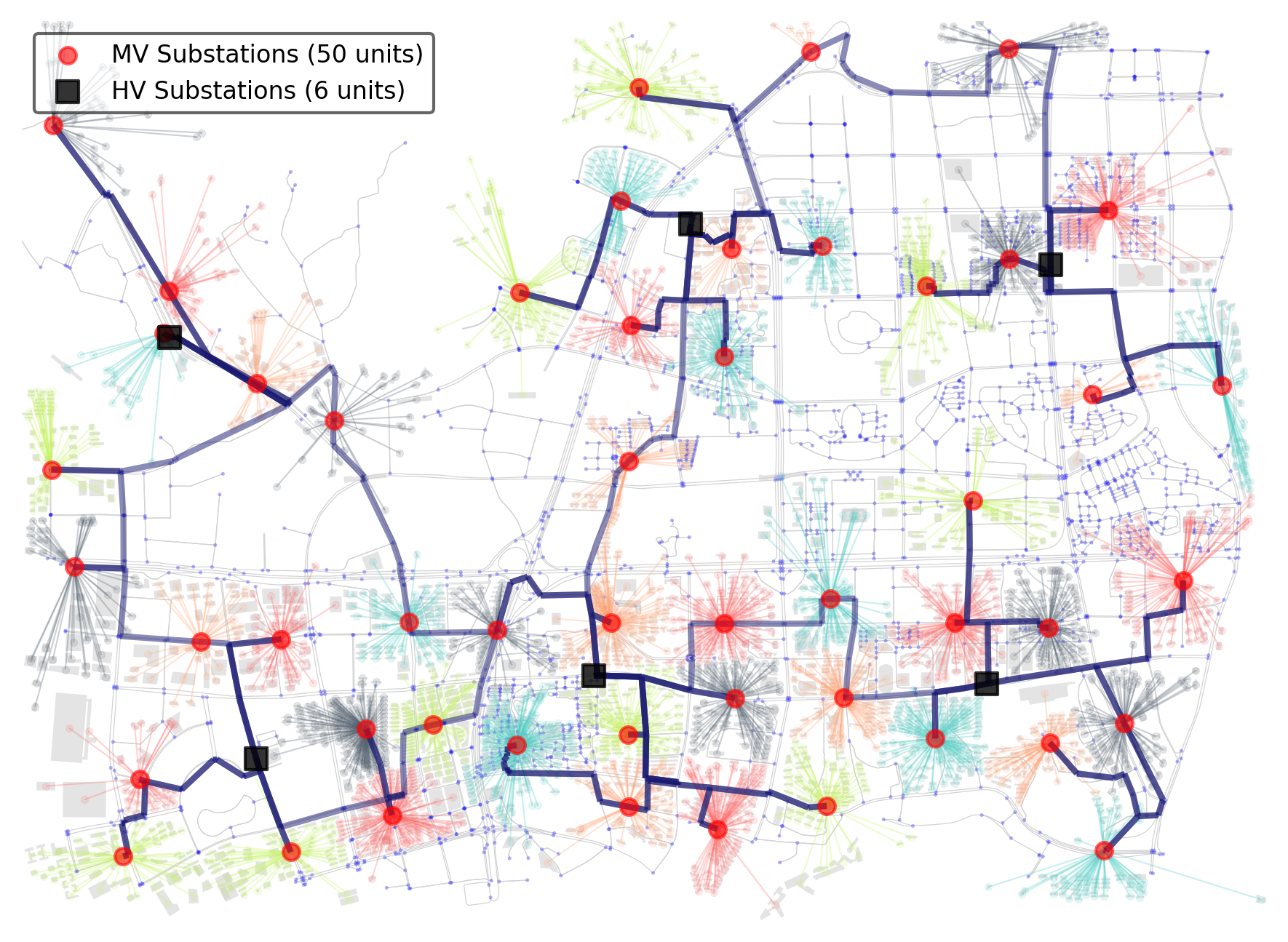}
    \\ \textbf{(b)} Solution for Case6 (Wangcheng)
  \end{minipage}
  \hfill
  \begin{minipage}{0.27\linewidth}
    \centering
    \includegraphics[width=\linewidth]{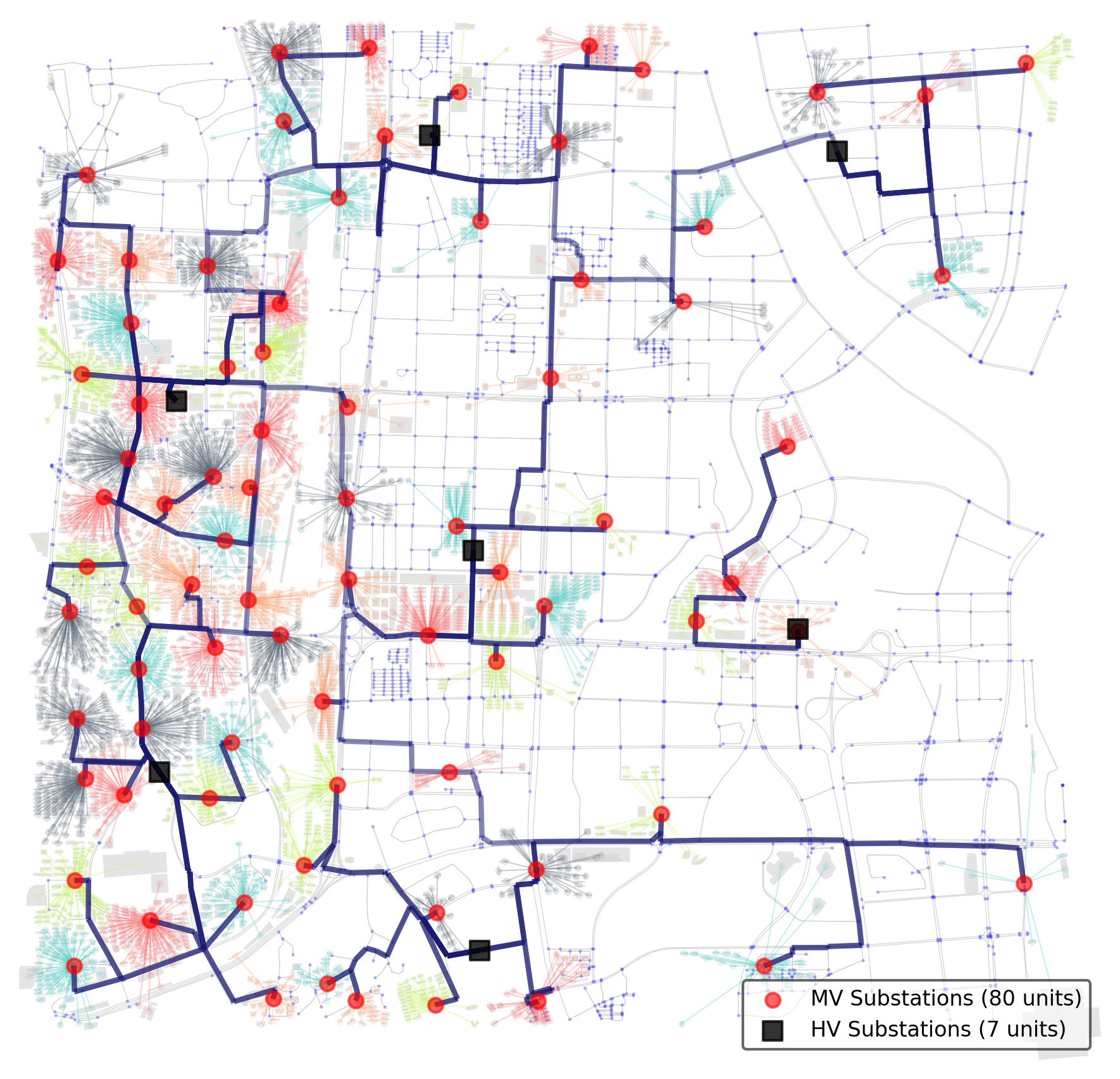}
    \\ \textbf{(c)} Solution for Case7 (East-Changsha)
  \end{minipage}
  
  \caption{Visualization of L-VNS solutions: colored dots denote residential demand points linked to MV substations (red nodes) via thin colored lines; black squares represent HV substations; dark blue lines depict cable routing between substations, with opacity indicating co-routed parallel cables.}
  \label{fig:gis_solutions}
\end{figure*}

Across all three real-world instances, L-VNS consistently achieves the lowest construction costs, with average RPD of 1.67\%, 1.16\%, and 0.92\% relative to SANS-Kriging, SANS-MLP, and GVNS respectively. Despite facing road network constraints and complex terrain dependencies, L-VNS demonstrates robust generalization from synthetic benchmarks to real urban environments. The significance tests confirm that L-VNS significantly outperforms all three baselines on all instances, underscoring its practical utility for urban distribution network planning. Fig. \fref{fig:gis_solutions} further visualizes the solutions obtained by L-VNS on OpenStreetMap.

\section{Further Discussions}
\label{discussions}

From a methodological viewpoint, the key difficulty of urban underground cable routing lies in the strong bilevel coupling between connectivity decisions and road-constrained path realization. Our L-VNS addresses this coupling by (i) generating a high-quality connectivity ``anchor'' via an auxiliary upper-level task, (ii) exploiting locality in neighborhood transitions to avoid full lower-level re-optimization, and (iii) explicitly modeling trench-sharing economies through a modified $\mathbb{A}^*$ re-planning operator. These design choices suggest a general principle for solving similar bilevel combinatorial problems: incorporate the most informative lower-level structure into guided sampling, while maintaining incremental feasibility to control the evaluation cost.

For complex bilevel optimization problems, many existing studies place greater emphasis on approximating the lower-level optimization (e.g., via surrogate or proxy models) or on coordinating multi-fidelity optimization. The effectiveness of the proposed auxiliary initialization mechanism and the DRL-based sampling strategy in this work indicates that reducing the search space at the upper level, thereby avoiding unnecessary lower-level evaluations, can be also beneficial for improving overall bilevel efficiency.

Whereas showing good performance, the introduction of DRL entails a non-negligible offline training time, while the performance gain remains moderate: compared with GVNS without DRL, the proposed method reduces the cost by approximately $2\%$. Although this level of improvement is practical for this planning setting and can correspond to savings on the order of millions of CNY, it is important to recognize that learning with DRL remains challenging when the decision space is large.

\section{Conclusions}
\label{conclusion}

This paper addresses the urban cable routing problem by formulating it as a bilevel optimization framework that explicitly considers road-constrained routing and trench-sharing cost savings. We propose a learning-assisted variable neighborhood search (L-VNS) algorithm that hierarchically optimizes substation connectivity and detailed cable routes. The framework integrates four key components: auxiliary task-based initialization for high-quality starting solutions, variable neighborhood search for upper-level connectivity optimization with multiple search operators, multi-agent deep reinforcement learning for probabilistic neighborhood sampling, and incremental lower-level re-planning with trench-sharing-aware $\mathbb{A}^*$ routing. These mechanisms collectively enable efficient navigation of the exponentially large bilevel search space. Computational experiments on 12 benchmark instances and 3 GIS-derived instances demonstrate that L-VNS significantly outperforms competitive baselines, reducing average costs by 0.92–73.72\% relative to representative methods. Ablation studies and sensitivity analyses further validate the effectiveness and robustness of the algorithm.

Despite these advances, fully leveraging deep reinforcement learning to augment neighborhood search remains challenging. Future work will focus on improving agent learning efficiency and generalization across diverse problem instances, as well as extending the framework to multi-objective optimization with reliability considerations under stochastic failures.

\section*{CRediT authorship contribution statement}
\textbf{Wei Liu}: Writing – original draft, Investigation, Visualization, Validation, Supervision, Methodology. \textbf{Rui Wang}: Writing – review \& editing, Supervision, Funding acquisition, Conceptualization. \textbf{Chenhui Lin}: Writing – review \& editing, Supervision, Project administration, Formal analysis. \textbf{Kaiwen Li}: Writing – original draft, Validation, Conceptualization. \textbf{Wenhua Li}: Funding acquisition, Supervision. \textbf{Tao Zhang}: Writing – review \& editing, Investigation, Funding acquisition.

\section*{Declaration of competing interest} 
The authors declare that they have no known competing financial interests or personal relationships that could have appeared to influence the work reported in this paper.

\section*{Acknowledgments} 
This work was supported by the financial support provided by the National Natural Science Foundation of China (62550020, 72421002, 62303476, 62503488, 52307101), and the Hunan Provincial Innovation Foundation for Postgraduate (CX20240104).

\section*{Code availability} 
The source code for this study will soon be available open-source \footnote{https://github.com/Will-iam-L/L-MVNS-for-urban-cable-routing}.

\bibliographystyle{elsarticle-num} 
\bibliography{test}

\end{document}